\documentclass{article}
\usepackage{fullpage}
\usepackage[utf8]{inputenc} 
\usepackage[T1]{fontenc}    
\usepackage{url}            
\usepackage{booktabs}       
\usepackage{amsfonts}       
\usepackage{nicefrac}       
\usepackage{microtype}      
\usepackage{soul}
\usepackage{graphicx}
\usepackage{amsmath}
\usepackage{outlines}
\usepackage{xurl}

\usepackage[colorlinks=true,
citecolor=blue,
filecolor=black,
linkcolor=blue,
urlcolor=blue]{hyperref}

\usepackage{amsmath,amsfonts,bm}

\usepackage{xcolor}
\usepackage{colortbl}

\def\eqref#1{equation~\ref{#1}}

\DeclareMathAlphabet{\mathsfit}{\encodingdefault}{\sfdefault}{m}{sl}
\SetMathAlphabet{\mathsfit}{bold}{\encodingdefault}{\sfdefault}{bx}{n}

\usepackage{multirow}
\usepackage{paralist}

\usepackage{multicol}
\usepackage{diagbox}

\usepackage[ruled,noend]{algorithm2e}

\SetCommentSty{mycommfont}

\usepackage{here}

\usepackage{amsmath,amssymb,amsfonts,amsbsy,amsfonts,latexsym}
\usepackage{makecell}
\usepackage{xcolor}
\usepackage{colortbl}

\usepackage{tabularx,colortbl,xcolor}
\usepackage[normalem]{ulem}
\useunder{\uline}{\ul}{}

\usepackage{enumitem}

\usepackage{xparse}

\SetKwInput{KwInput}{Input}
\SetKwInput{KwRequire}{Require}

\usepackage{longtable}

\NewDocumentCommand{\var}{O{s} m O{}}{%
  \ensuremath{#1_{#2}^{#3}}
}
\usepackage{siunitx}

\newcommand{\commentout}[1]{}

\definecolor{light-gray}{gray}{0.80}

\newcommand\appref{Appendix~\ref}

\newcommand\fref{Figure~\ref}
\newcommand\tref{Table~\ref}
\newcommand\sref{Section~\ref}

\usepackage{amsthm}

\usepackage{amsthm}

\newtheorem{mytheorem}{Theorem}
\newtheorem{assumption}[mytheorem]{Assumption}
\newtheorem{lemma}[mytheorem]{Lemma}
\newtheorem{remk}[mytheorem]{Remark}

\usepackage{subfig}
\usepackage{tcolorbox}
\usepackage{wrapfig,lipsum,booktabs}
\usepackage{pifont}
\newcommand{\cmark}{\ding{51}}%
\newcommand{\xmark}{\ding{55}}%

\newcommand{\ptq}{PTQ\xspace}
\newcommand{\lorc}{LoRC\xspace}

\newcommand{\ppl}{PPL\xspace}
\newcommand{\llm}{LLM\xspace}
\newcommand{\llms}{LLMs\xspace}

\newcommand{\opt}{OPT\xspace}
\newcommand{\bloom}{BLOOM\xspace}
\newcommand{\rtn}{RTN\xspace}
\newcommand{\gptq}{GPTQ\xspace}
\newcommand{\zqglobal}{ZQ-Global\xspace}
\newcommand{\zqlocal}{ZQ-Local\xspace}
\newcommand{\zqglobalstar}{ZQ-Global$^*$\xspace}
\newcommand{\zqlocalstar}{ZQ-Local$^*$\xspace}
\newcommand{\sym}{$^\text{sym}$\xspace}
\newcommand{\asym}{$^\text{asym}$\xspace}

\newcommand{\classone}{\textbf{\textit{Class}}-1\xspace}
\newcommand{\classtwo}{\textbf{\textit{Class}}-2\xspace}
\newcommand{\classthree}{\textbf{\textit{Class}}-3\xspace}

\usepackage{chngcntr}
\usepackage{adjustbox}

\newcommand*\samethanks[1][\value{footnote}]{\footnotemark[#1]}

\begin{document}

\title{
ZeroQuant-V2: Exploring Post-training Quantization in LLMs from Comprehensive Study to Low Rank Compensation
}

\author{
Zhewei Yao\thanks{Equal Contribution. Code will be released  as a part of \url{https://github.com/microsoft/DeepSpeed}}, Xiaoxia Wu$^*$,  Cheng Li,  Stephen Youn,  Yuxiong He
\\  Microsoft \\ 
{\tt \small\{zheweiyao,  xiaoxiawu, chengli1, stephen.youn, yuxhe\}@microsoft.com}
}

\date{}
\maketitle
\begin{abstract}

Post-training quantization (PTQ) has emerged as a promising technique for mitigating memory consumption and computational costs in large language models (LLMs). However, a systematic examination of various quantization schemes, model families, and quantization bit precision has been absent from the literature. In this paper, we conduct a comprehensive analysis of these factors by investigating the effects of PTQ on weight-only, activation-only, and weight-and-activation quantization using diverse methods such as round-to-nearest (RTN), GPTQ, ZeroQuant, and their variants. We apply these methods to two distinct model families with parameters ranging from 125M to 176B. Our contributions include: (1) a sensitivity analysis revealing that activation quantization is generally more susceptible to weight quantization, with smaller models often outperforming larger models in terms of activation quantization; (2) an evaluation and comparison of existing PTQ methods to optimize model size reduction while minimizing the impact on accuracy, revealing that none of the current methods can achieve the original model quality for quantization with either INT4-weight or INT4-weight-and-INT8-activation; (3) based on these insights, we propose an optimized method called Low-Rank Compensation (LoRC), which employs low-rank matrices to enhance model quality recovery with a minimal increase in model size.
\end{abstract}
\section{Introduction}
\label{sec:intro}

Large language models (LLMs) like Codex~\cite{copilot} and ChatGPT~\cite{chatgpt} have demonstrated breakthrough performance across various benchmarks, such as natural language understanding and generation, and are now integrated into everyday applications. However, efficiently serving LLMs has become a pressing concern due to their significant memory consumption and computational demands. Unlike classification or diffusion models, LLMs present unique challenges, as they involve two distinct phases: prompt and generation. The prompt phase is primarily compute-bound, while the generation phase, with low batch size and KV cache, is mainly memory-bound~\cite{pope2022efficiently}.

As the progression of hardware bandwidth lags behind that of computational demand \cite{gholami2020ai}, the resource demands of extra-large models such as MT-NLG-530B~\cite{smith2022using}—which necessitates the deployment of multiple nodes for operation—escalate, adding to the complexities of cross-node communication. This has emphasized the urgency to curtail both the size and computational expense of Large Language Models (LLMs). An increasingly effective solution to these issues is post-training quantization (PTQ). This method aids in the reduction of training prerequisites while simultaneously lowering the bit precision of weights and activations to either INT4 or INT8.

\begin{figure}
\centering
\includegraphics[width=0.49\textwidth]{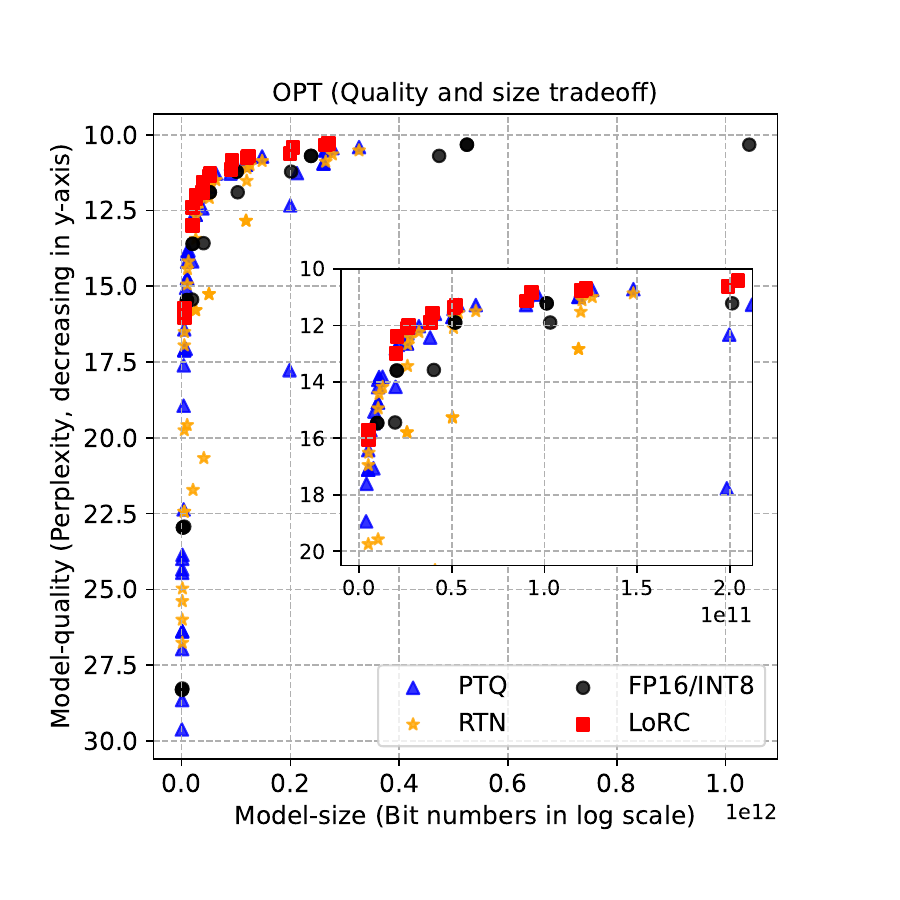}
\includegraphics[width=0.49\textwidth]{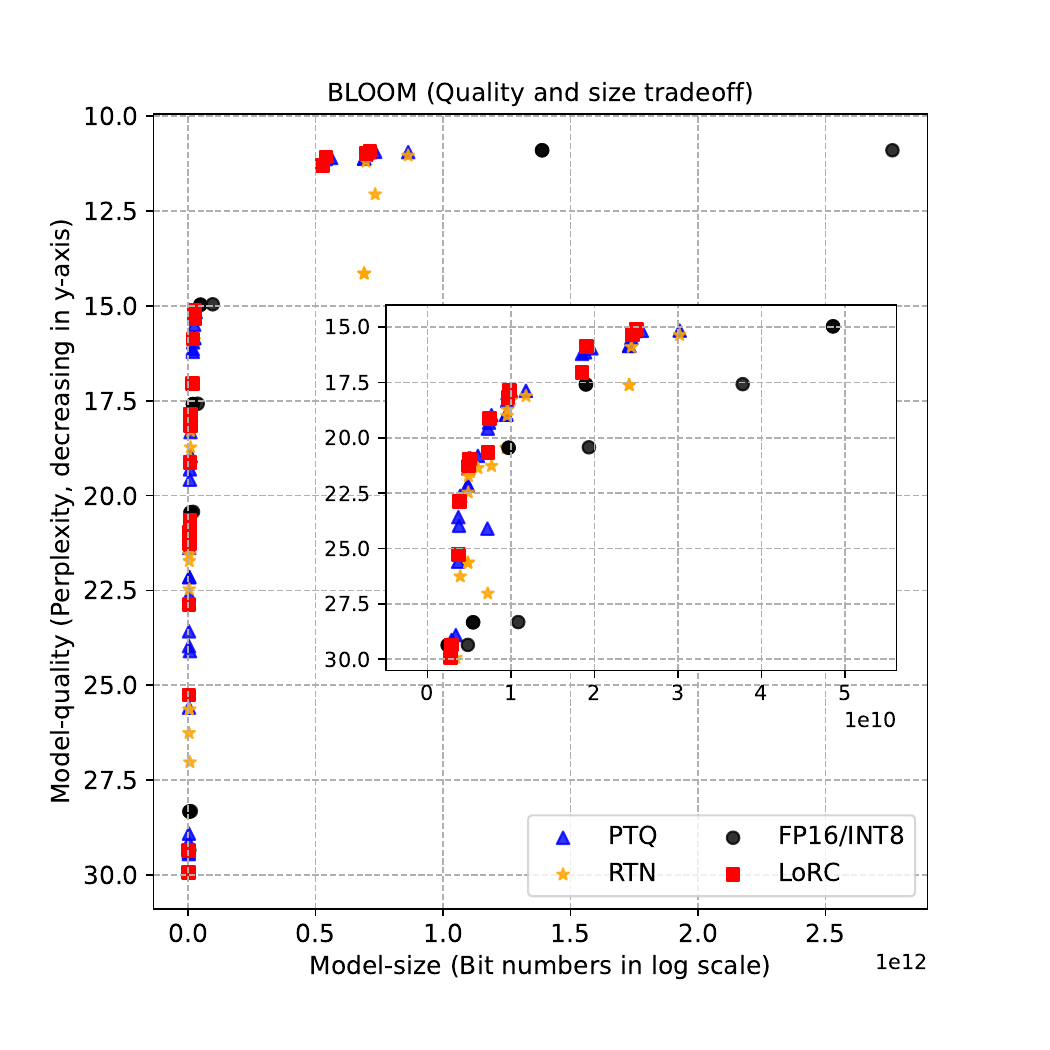}
\vspace{-0.42cm}
\caption{The model size and quality trade-off of different quantization methods on models from \opt and \bloom families.
Here \ptq (with fine-grained quantization) represents the method from~\cite{yao2022zeroquant,frantar2022gptq},
\rtn means the naive round-to-nearest baseline (with fine-grained quantization as well),
and FP16/INT8 is used as the no-accuracy-loss baseline. \lorc is our proposed method that works seamless with \ptq.
Note that we drop all diverged points for better visualization.
For all detailed numbers, please see~\appref{sec:full_tables_results}.}
\label{fig:main_figure}
\vspace{-0.5cm}
\end{figure}

While the effectiveness of post-training quantization (PTQ) has been underscored in a number of recent studies~\cite{yao2022zeroquant,frantar2022gptq,xiao2022smoothquant,dettmers2022case}, a comprehensive, systematic investigation into several key dimensions of this technique remains to be undertaken. Specifically, the extant literature falls short in providing thorough coverage of the functionality of various PTQ methods or the sensitivity of disparate models. Moreover, despite current quantization methods demonstrating promising results in the reduction of model sizes, the question persists as to whether these methods are achieving their optimal potential in minimizing Large Language Models (LLMs) sizes.

With these observations in mind, our study sets forth to address two salient questions: (1) When subjected to quantization, do LLMs of varying sizes and pretraining data exhibit similar behavior? (2) Are existing quantization methods truly leveraging their full potential in reducing the sizes of LLMs?

\paragraph{Contribution.} To elucidate these queries, we undertake an exhaustive examination of the impact of PTQ on weight-only, activation-only, and combined weight-and-activation quantization. This investigation incorporates a range of PTQ methods, including round-to-nearest (\rtn), GPTQ~\cite{frantar2022gptq}, ZeroQuant~\cite{yao2022zeroquant}, and their respective variants. To broaden the scope of our analysis, we focus on two distinct model families, OPT~\cite{zhang2022opt} and BLOOM~\cite{scao2022bloom}, spanning model sizes from 125M to a massive 176B. Our code will be made available for reproduction. In summary, we make the following contributions:

(1) We provide a thorough \textbf{sensitivity analysis} 
to demonstrate that a) Activation quantization is generally more sensitive to weight quantization; Smaller models usually have better activation quantization performance than the relative larger model. b) Different model families show different INT8 activation quantization behaviors; Particularly for large models, \bloom-176B has  small accuracy drops (about 1 perplexity or PPL) but \opt-30B and -66B experience worse performance. 

(2) We carry out a detailed evaluation and comparison of current PTQ methods, utilizing optimal configurations to maximize model size reduction while minimizing accuracy impact. 
We found that the current existing method can barely achieve less than 0.1 \ppl points degradation for quantization with either INT4-weight or INT4-weight-and-INT8-activation (W4A8). To recover the 0.1 \ppl, we strive to push the boundaries of employing \textbf{fine-grained quantization} (FGQ) techniques. We observe FGQ is able to recovered points degradation of <0.1 \ppl  for large models (>13B) for INT4 weight quantization, but there are still non-negligible model quality drops.

(3) Based on the above understanding, we further optimize existing methods and introduce a technique called \textbf{Lo}w \textbf{R}ank \textbf{C}ompensation (LoRC), which employs low-rank matrix factorization on the quantization error matrix. Complementary to FGQ, LoRC plays a crucial role in enhancing the full model quality recovery, while there is little increase of the model size.

In~\fref{fig:main_figure}, we provide model size and quality trade-offs for both \opt and \bloom families. 
As can be seen, using \lorc on top of \ptq  methods from~\cite{yao2022zeroquant,frantar2022gptq} and fine-grained quantization, we set a new quantization Pareto frontier for \llms.   
Meanwhile, we recommend the following setting for quantizing \llms with LoRC (Note that activation quantization should be only applied if necessary): 
(1) For larger models (>10B), fine-grained (block size 64--256) 4-bit weight quantization plus 8-bit activation quantization (block size 64--256) with \ptq  can be used for real deployment;
(2) For middle-size models (<10B and >1B), per-row INT8 quantization plus fine-grained (block size 64--256) INT8 activation quantization can be used with \ptq  from~\cite{frantar2022gptq,yao2022zeroquant};
(3) For smaller models (<1B),  per-row W8A8 (INT8 weight and INT8 activation) \rtn is enough based on~\cite{yao2022zeroquant}.

\section{Related Work}
\label{sec:related_work}
Different quantization methods~\cite{shen2020q,zafrir2019q8bert,fan2020training,zhang2020ternarybert,bai2020binarybert,esser2019learned,tao2022compression,kim2021bert} for transformer-based models~\cite{vaswani2017attention} have been explored for a while.
However, most of those works need quantization-aware finetuning or even expensive quantization-aware knowledge distillation~\cite{hinton2015distilling}.
Due to the cost of training/finetuning \llms \cite{polino2018model,jiao2019tinybert,tao2022compression,wu2022extreme,wu2023understanding}, it is a challenge for practitioners/researchers to do finetuning/distillation on those \llms, particularly for models like GPT-3-175B~\cite{brown2020language} and \bloom-176B~\cite{scao2022bloom}. 

Post-training quantization (\ptq)~\cite{zadeh2020gobo,bondarenko2021understanding} is an alternative way to quantize the model with no/minimal finetuning requirement.
Along this line, several recent works focus on \llms (beyond the million-parameter scale).
\cite{yao2022zeroquant} proposes vector-based INT8 quantization with layer-by-layer knowledge distillation to overcome the training cost and quantization error introduced by \llms. 
\cite{dettmers2022llm} uses similar vector-based INT8 quantization weight plus mixed-precision (INT8/FP16) quantization for activation to overcome the sensitivity of activation quantization. 
However, the inference speed of \cite{dettmers2022llm} is generally even slower than FP16 baseline~\cite{bloom_inference} due to the difficulty of implementing mixed-precision calculation within a single tensor.
More recently, \cite{frantar2022gptq} extends OBQ~\cite{frantar2022optimal,hassibi1993second, lecun1990optimal}    on \llms for INT4 weight-only quantization and shows great efficiency on quantization and latency,
and \cite{xiao2022smoothquant} shows the outliers from activations can be smoothed out by migrating the quantization difficulty from activations to its associated weights. 
However, \cite{xiao2022smoothquant} can only work for W8A8 quantization as lower weight precision (INT4) itself already leads to significant accuracy degradation, and the accuracy drop is larger than 0.1 \ppl points, which as discussed in the later section is sub-optimal.
\cite{dettmers2022case} shows the scaling law of weight-only quantization with the simplest round-to-nearest baseline, but it does not consider the weight-and-activation quantization and/or the above \ptq optimization methods.
As can be seen from~\fref{fig:main_figure}, by using \ptq optimization methods, the model quality can be significantly improved. 
Please also see~\appref{sec:full_tables_results} for more detailed numbers.

Different than existing works, our paper extensively tests the effect of (1) different quantization schemes, e.g., symmetric and asymmetric quantization, (2) different \ptq methods, e.g.,~\cite{yao2022zeroquant,frantar2022gptq}, (3) different model families, e.g.,~\cite{scao2022bloom,zhang2022opt}, (4) different quantization coverage, e.g., weight-only and weight-and-activation quantization, and (5) other discussions, e.g., the effect of quantization granularity. 
As such, we provide a much more comprehensive understanding of post-training quantization for large language models compared to the previous works.
\section{Would different model families behave similarly on quantization?}
\label{sec:ptq_challenge}

There are mainly two categories of \ptq for \llms, i.e., weight-only quantization~\cite{frantar2022gptq} and weight-and-activation quantization~\cite{dettmers2022llm,yao2022zeroquant,xiao2022smoothquant}. 
In the latter, it is uniformly observed across all studies that activation quantization demonstrates greater sensitivity than weight quantization. However, prior research tends to concentrate on a single (family) model to emphasize the necessity of their proposed quantization technique. A comprehensive and systematic evaluation of this PTQ methodology, particularly the sensitivity of weight/activation quantization for varying model sizes and distinct model families, has yet to be undertaken. Hence, we conduct an examination on both the OPT~\cite{zhang2022opt} and BLOOM~\cite{scao2022bloom} families to elucidate the quantization sensitivity of weight and activation.

\begin{wraptable}{r}{8cm}
\caption{
Classification of quantization sensitivity (or quantization loss). The sensitivity increases from \classone to \classthree.
}\centering
\label{tab:quantization-loss-table}
\begin{adjustbox}{width=1.0\linewidth}
\centering
\begin{tabular}{lcccccccccccccc }
\toprule
Class     & \classone & \classtwo & \classthree \\
\midrule
\ppl Degradation & $\le$0.1 & $>$0.1 \& $\le$0.5 & $>$0.5 \\ 
\bottomrule
\end{tabular}
\end{adjustbox}
\vspace{-0.2cm}
\end{wraptable}
\textbf{Sensitivity setting.} 
We use the zero-shot validation perplexity (\ppl) differential on three datasets, namely, Wikitext-2~\cite{merity2016pointer}, PTB~\cite{marcinkiewicz1994building}, and C4~\cite{colin2019t5}, before and after the quantization of these LLMs to illustrate their sensitivity, as \ppl is significantly correlated to zero-shot/few-shot accuracy measurement~\cite{dettmers2022case}. Specifically, a higher \ppl drop indicates enhanced quantization sensitivity. For simplicity, we also categorize quantization sensitivity (or quantization loss) into three different classes as depicted in~\tref{tab:quantization-loss-table}. Notably,
the threshold is chosen because when the model size approximately doubles (e.g., 13B vs. 30B, and 30B vs. 66B), the \ppl improvement is about 0.5 (see~\tref{tab:ptq_challenge_opt_average_in_maintext}).  The sensitivity (or loss) incrementally increases as the class number ascends. From a practical standpoint, we favor lower quantization sensitivity (accuracy loss), making \classone the optimal-loss post-training quantization. 

We employ both symmetric and asymmetric quantization to gauge the quantization sensitivity and highlight the advantage of asymmetric quantization. Particularly, we implement per-row quantization~\cite{frantar2022gptq} for weight quantization and per-token quantization for activation~\cite{yao2022zeroquant}.

\textbf{Robustness of Weight-only Quantization for Large Models.} 
The results of weight-only quantization in \opt and \bloom models are summarized in~\tref{tab:ptq_challenge_opt_average_in_maintext}. INT8 weight-only quantization, either symmetric or asymmetric, results in negligible accuracy loss (less than 0.05, i.e., \classone). Consequently, for tasks oriented towards generation, FP16 weight can simply be replaced with INT8 weight to reduce memory usage. For INT4 quantization, the asymmetric method outperforms the symmetric approach in accuracy, attributable to its superior utilization of the quantization range. Interestingly, larger models exhibit better tolerance to low-precision quantization (i.e., INT4) than smaller models, with a few exceptions such as \opt-66B.\footnote{\cite{frantar2022gptq} discovered that \opt-66B has a high proportion of dead neurons in the early layers, which might influence the compression capability. We also identify another potential reason: the Layer Norm of the \opt-family is not well trained (except \opt-350M), with the weight and the bias being all 1's and 0's, respectively.} Particularly, \bloom-176B shows \ppl degradation (around 0.3 points) in \classtwo, which could explain why the large GLM-130B~\cite{zeng2022glm} can operate with INT4 weight-only quantization out of the box with acceptable accuracy impact.

\begin{table}[H]
\caption{
Average \ppl of \opt and \bloom (BLM). 
See~\tref{tab:ptq_challenge_opt_full_in_appendix} for all results.
}\centering
\label{tab:ptq_challenge_opt_average_in_maintext}
\begin{adjustbox}{width=0.999\linewidth}
\centering
\begin{tabular}{lcccccccccccccc }
\toprule
Precision     &  OPT-6.7b	& OPT-13b &OPT-30b	& OPT-66b   & BLM-1.7b  & BLM-3b & BLM-7.1b & BLM-176b\\
\midrule
W16-A16     & 11.90 & 11.22   & 10.70   & 10.33   & 20.43 & 17.58 & 14.96 & 10.90 \\\midrule
W8\sym-A16  & 11.90 & 11.22   & 10.70   & 10.33   & 20.43 & 17.59 & 14.97 & 10.90 \\
W8\asym-A16 & 11.90 & 11.22   & 10.70   & 10.33   & 20.45 & 17.59 & 14.97 & 10.90 \\ \midrule
W4\sym-A16  & 14.36 & 12.73   & 11.77   & 97.05   & 23.18 & 19.36 & 16.27 & 11.28 \\
W4\asym-A16 & 13.44 & 12.09   & 11.52   & 31.52   & 22.47 & 19.01 & 15.90  & 11.20  \\ \midrule
W16-A8\sym  & 26.04 & 3171.49 & 2048.21 & 2638.09 & 20.68 & 17.73 & 15.28 & 12.10  \\
W16-A8\asym & 12.62 & 15.36   & 23.57   & 561.35  & 20.52 & 17.65 & 15.14 & 11.62\\
\bottomrule
\end{tabular}
\end{adjustbox}
\end{table}

\textbf{Challenge Encountered in Activation Quantization for Large Models.}
Activation quantization has consistently proven more difficult than weight quantization~\cite{yao2022zeroquant,dettmers2022llm}, as illustrated in \tref{tab:ptq_challenge_opt_average_in_maintext}. When compared to weight-only quantization, activation-only quantization indicates that asymmetric quantization can significantly improved performance over symmetric quantization. Moreover, contrary to weight-only quantization, smaller models typically exhibit better tolerance to activation quantization, as their hidden dimension is smaller and the activation dynamic range is also narrower than larger models~\cite{yao2022zeroquant}. It should be noted that for models larger than 10B, all fall into \classthree, indicating a degradation of more than 0.5 \ppl points. 


The last two rows of \tref{tab:ptq_challenge_opt_average_in_maintext} show that different model families exhibit significantly different behaviors. \bloom does not exhibit divergence issues even up to a model size of 176B, whereas \opt displays very poor performance from a model size of 6.7B (larger models with INT8 activation have even worse \ppl). This could again be attributed to the Layer Norm issue within the \opt-family\samethanks.


\begin{tcolorbox}
\textbf{Findings 1 on Sensitivity Analysis.} \textbf{(1)} INT8 weight-only quantization can serve as a standard method for reducing memory costs in \llms, with negligible degradation in accuracy. \textbf{(2)} INT4 weight-only quantization for small models results in substantial accuracy degradation (\classthree), but this effect lessens as the model size increases (\classtwo).
\textbf{(3)} Contrary to (2), INT8 activation results in minimal accuracy drops for small models (\classone) but larger models exhibit greater drops (\classthree). \textbf{(4)} With INT8 activation, \bloom shows no divergence issues up to a model size of 176B, whereas \opt performs poorly from $\geq$ 6.7B model sizes.
\end{tcolorbox}
\section{Are existing quantization methods optimally harnessing the potential to minimize LLMs sizes?}
\label{sec:evaluation_of_existing_methods}
Numerous lightweight optimization-based methods have been proposed, which update the model weights during quantization. These methods such as \cite{yao2022zeroquant,frantar2022gptq,xiao2022smoothquant}, unlike quantization-aware training, only require a small portion of the training data and a limited training time. Particularly, GPTQ~\cite{frantar2022gptq} and ZeroQuant~\cite{yao2022zeroquant}, have proven to be effective and efficient in terms of GPU resources, time cost, and data usage for INT4 weight quantization.\footnote{We tested the method proposed by \cite{xiao2022smoothquant} but did not find it better than others for INT4 weight quantization.} In this work, we focus on the variants of GPTQ and ZeroQuant as well as the most straightforward baseline, round-to-nearest neighborhood (RTN).

\textbf{\rtn} directly applies \ptq on the trained data and follows the procedure detailed in Section~\ref{sec:background_of_quantization} to perform the quantization. Specifically, for symmetric quantization, we set $S=max(abs(x))$ and $Z=0$; for asymmetric quantization, we set $S=max(x)-min(x)$ and $Z=min(x)$.

\textbf{\gptq} extends the OBQ~\cite{frantar2022optimal}. It  tries to optimize the following non-linear least square problem,
    $\min_{\hat W} \|Wx - \hat Wx\|_2^2$
where $W$ is the weight, $x$ is the activation, and $\hat W$ is a quantized weight. GPTQ employs second-order methods to obtain a closed-form solution. In addition, the quantization for each weight matrix is performed column-/row-wisely and the quantization errors from previous columns will be passed to those columns not yet quantized. See\cite{frantar2022optimal,frantar2022gptq} for more details.

\textbf{\zqglobal} is the original method proposed in~\cite{yao2022zeroquant}, where authors treat each layer as a small neural network (a.k.a., subnetwork) and use the FP16 subnetwork as the teacher model to distill the quantized one with a few hundred iterations, i.e., $ \min_{\hat \theta} |f_{\theta}(x) - f_{\hat\theta}(x)|2^2,$ where $\theta$ is a set of weights, $\hat \theta$ is the quantized version, $f{\theta}$ is the subnetwork with parameters $\theta$, and $x$ is the input. Thus, it can significantly reduce the GPU resource requirement and time cost.


\textbf{\zqlocal} is an extension mode of \zqglobal for further GPU requirement reduction and training cost reduction. 
Particularly, instead of using each transformer layer as the subnetwork, we treat each linear layer as the subnetwork. 
This method can be viewed as an iterative first-order optimization method (e.g., SGD) to solve~ $\min_{\hat W} \|Wx - \hat Wx\|_2^2$. 

\textbf{Experimental Setup.} We compare the four methods mentioned above on weight-only and weight-and-activation quantization. As weight quantization is always static (i.e., it does not change during inference), there is virtually no system performance difference between symmetric and asymmetric quantization.\footnote{The bias term (a.k.a., the zero point) can be simply fused into the previous activation quantization kernel~\cite{yao2022zeroquant}.} We use asymmetric quantization for better accuracy, and the conclusions would hold similarly for symmetric quantization. For parameters used for \gptq, \zqlocal, and \zqglobal, please refer to Appendix~\ref{sec:hyperparameter_used_in_exisiting_method_evaluation}. An interesting finding for ZeroQuant is that the hyperparameters (e.g., learning rate and its scheduler) provided in the original work~\cite{yao2022zeroquant} are sub-optimal. In this work, we find the best configurations for \zqlocal and \zqglobal and denote them as \zqlocalstar and \zqglobalstar, respectively, with the best tuned results. To ensure consistent and comparable results, we set a fixed random seed for our experiments. In the context of post-training quantization, varying the random seed has minimal impact on the final results, as indicated in more detail in \tref{tab:mean}.

\textbf{Evaluation of Weight-only Quantization.}\label{sec:weight_only_quantization_existing_method} The results from weight-only quantization using OPT and Bloom are presented in Table~\ref{tab:weight_only_quantization_opt_existing_method_average_in_main_text}. The findings indicate that the larger models tend to be less sensitive to INT4 weight-only quantization. This observation holds true across all methods (\rtn, \gptq, \zqlocalstar, and \zqglobalstar) with the exception of OPT-66B, which shows greater degradation than OPT-30B. It is noteworthy that light-weight optimization-based methods significantly outperform the \rtn baseline in terms of accuracy. For instance, these methods substantially reduce the degradation in perplexity of OPT-30B/66B compared to baseline. Most quantized models with parameters greater than 6.7B fall under Class II, indicating their potential for real-world applications. For instance, the quality of INT4 OPT-30B (66B) is superior to that of INT8 OPT-13B (30B).


Among the optimization-based methods, \zqglobalstar  generally performs better on smaller models (those with fewer than 1B parameters), while \gptq excels on larger models. \zqlocalstar does not outperform \gptq or \zqglobalstar -— a reasonable outcome given that  \gptq  employs a closed-form solution to solve the non-linear quadratic problem and \zqglobalstar optimizes a larger subnetwork. The inferior performance of \zqglobalstar compared to \gptq for larger models is unexpected since \zqglobalstar  optimizes an entire transformer layer while \gptq only optimizes a single linear layer. A plausible explanation is that larger models are more sensitive to weight updates, necessitating more advanced fine-tuning methods.


\begin{table}[t] 
\vspace{-0.5cm}
\caption{
The evaluation results of different \ptq methods on \opt and \bloom (BLM) with asymmmetric quantization on weight or (and) activation.
See more details in ~\tref{tab:weight_only_quantization_opt_existing_method_full_in_appendix} and~\tref{tab:weight_only_quantization_bloom_existing_method_full_in_appendix}.
}\centering
\label{tab:weight_only_quantization_opt_existing_method_average_in_main_text}
\begin{adjustbox}{width=0.999\linewidth}
\centering
\begin{tabular}{lcccccccccccccc }
\toprule
Precision &    Method               & OPT-6.7b & OPT-13b & OPT-30b & OPT-66b & BLM-1.7b & BLM-3b & BLM-7.1b & BLM-176b \\\midrule
\multicolumn{2}{l}{W16A16}                      & 11.90     & 11.22   & 10.70    & 10.33   & 20.43    & 17.58  & 14.96    & 10.90     \\\midrule
\multirow{4}{*}{W4A16} &\rtn          & 13.44    & 12.09   & 11.52   & 31.52   & 22.47    & 19.01  & 15.90     & 11.20     \\
&\gptq         & 12.28    & 11.42   & 10.78   & 10.52   & 21.58    & 18.33  & 15.50     & 11.02    \\
&\zqlocalstar  & 12.46    & 11.64   & 11.05   & 10.79   & 21.70     & 18.50   & 15.55    & 11.11    \\
&\zqglobalstar & 12.38    & 11.62   & 11.04   & 10.68   & 21.38    & 18.33  & 15.52    & 11.05  \\ \midrule
\multirow{4}{*}{W4A8}
&\rtn          & 14.80     & 26.36   & 86.26   & 815.00     & 22.75    & 19.17  & 16.19    & 12.22    \\
&\gptq         & 13.88    & 17.28   & 20.71   & 648.69  & 21.71    & 18.44  & 15.75    & 11.86    \\
&\zqlocalstar  & 13.24    & 14.23   & 18.53   & 16.32   & 21.86    & 18.66  & 15.75    & 11.19    \\
&\zqglobalstar & 13.17    & 13.07   & 14.65   & 37.82   & 21.43    & 18.39  & 15.58    & 11.49   \\
\bottomrule
\end{tabular}
\end{adjustbox}
\vspace{-0.5cm}
\end{table}

\textbf{Evaluation of Weight and Activation Quantization.}
\label{sec:weightactivation_quantization_existing_method}
The evaluation results for existing methods using W4A8 quantization are presented in Table~\ref{tab:weight_only_quantization_opt_existing_method_average_in_main_text}. The three light-weight optimization-based methods  outperform \rtn significantly, underscoring their efficacy. However, all of the results fall into either  \classtwo or \classthree. This suggests that for certain applications, it might be more beneficial to use smaller models with fewer parameters rather than larger, quantized models.

Among quantization-based methods, \zqglobalstar and \zqlocalstar generally outperform GPTQ, which is anticipated given that GPTQ was originally designed for weight-only quantization. \zqglobalstar performs better than  \zqlocalstar in most cases except for the two largest models, OPT-66B and Bloom-176B, despite having larger trainable parameters in one step. This again signifies the need for a more suitable and advanced optimization method for large language models (LLMs).


\begin{tcolorbox}
\textbf{Finding 2 on Comparisons.}   \textbf{(1)} \gptq typically performs better for weight-only quantization, while ZeroQuant (including both \zqglobalstar and \zqlocalstar) yields superior results for weight and activation quantization. 
    \textbf{(2)} The tested optimization-based methods cannot achieve \classone quantization error for either INT4 weight-only or W4A8 quantization  with the exception of \gptq on OPT-30B with weight-only quantization.
\end{tcolorbox}

\subsection{Fine-grained Quantization and Its Evaluation}
With \ptq and row-wise quantization, achieving \classone quantization error is challenging for both weight-only and weight-and-activation quantization. Generally, utilizing a smaller model with INT8 weight is more advantageous than employing a model that is twice as large with INT4 weight.

One potential solution to this issue is the implementation of finer-grained quantization schemes~\cite{darvish2020pushing}, where every $k$ elements possess their own scaling factor and/or zero point. This approach can significantly reduce quantization error. In the extreme case, where every single element has its own scaling factor, the original FP16 number can be precisely recovered. Importantly, block-k quantization can be implemented on modern GPUs, one of the most prevalent deep learning architectures, since the compute unit (streaming multiprocessor) of GPUs processes tiles of data (e.g., 128 by 128 tiling size) for matrix computation.


Although fine-grained quantization can substantially narrow the gap between the quantized tensor and its floating-point counterpart, the application of \rtn still results in a non-trivial accuracy gap. Consequently, we build upon fine-grained quantization by employing existing optimization-based methods to further enhance accuracy. Specifically, we utilize \gptq and \zqglobal for all models and settings and apply \zqlocal to OPT-66B and Bloom-176B. For the hyperparameters used in \zqglobal and \zqlocal, we select the top three identified in Section~\ref{sec:evaluation_of_existing_methods} for all models, except for Bloom-176B, for which we only use the top-performing hyperparameter  to reduce training costs.
\label{sec:main_result_weightonly_quantization}
\paragraph{4-bit Weight Quantization.} We hereby present the W4A16 results for \opt and \bloom, as delineated in \tref{tab:opt-4bit-blocksize}, corresponding to an array of quantization block sizes. The performance sees a significant improvement with smaller block sizes compared to per-row quantization. The point of diminishing returns, however, varies for different model sizes. For example, smaller models (such as \opt-6.7B and \bloom-1.7b) continue to see substantial gains until the block size reduces to 32. In contrast, for larger models (those exceeding 10B, with \opt-66B as the exception), the benefits derived from smaller block sizes wane rapidly around block-256/512.  Most crucially, for models equal to or larger than 13B, a smaller quantization block size results in quantization error being classified under \classone, indicating virtually negligible degradation in accuracy.
\begin{table}[t]
\caption{
Results of \textbf{W4\asym-A16} quantization with various block-size out of the best result from optimization-based methods on \opt and \bloom (BLM). 
See~\tref{tab:opt-4bit-blocksize-full}  and~\tref{tab:bloom-4bit-blocksize-full} 
 for full results including \rtn. 
N/A means that the block size is not divisible by the hidden size.
}\centering
\label{tab:opt-4bit-blocksize}
\begin{adjustbox}{width=0.999\linewidth}
\centering
\begin{tabular}{lcccccccccccccc }
\toprule
Block-size  & OPT-6.7b & OPT-13b & OPT-30b & OPT-66b & BLM-1.7b & BLM-3b & BLM-7.1b & BLM-176b \\ \midrule 
W16A16   & 11.90     & 11.22   & 10.70    & 10.33   & 20.43    & 17.58  & 14.96    & 10.90     \\
Per-row   & 12.28    & 11.42   & 10.78   & 10.52   & 21.38    & 18.33  & 15.50     & 11.02    \\\midrule
1024      & 12.16    & 11.36   & 10.75   & 10.52   & 31.03    & N/A    & 15.24    & 10.96    \\
512       & 12.08    & 11.32   & 10.73   & 10.52   & 20.93    & 17.99  & 15.20     & 10.95    \\
256       & 12.05    & 11.28   & 10.74   & 10.50    & 20.95    & 17.97  & 15.18    & 10.95    \\
128       & 12.10     & 11.28   & 10.74   & 10.44   & 20.92    & 17.90   & 15.17    & 10.94    \\
32        & 12.03    & 11.28   & 10.72   & 10.41   & 20.82    & 17.88  & 15.16    & 10.95   \\
\bottomrule
\end{tabular}
\end{adjustbox}
\end{table}
\begin{table}[H]
\caption{
\opt W4\asym-A8 with various block-size out of the best result from \gptq, \zqlocal, and \zqglobal on \opt and \bloom (BLM). 
See~\tref{tab:opt-4bit8bit-blocksize-full} for full results including \rtn.
}\centering
\label{tab:opt-4bit8bit-blocksize}
\begin{adjustbox}{width=0.999\linewidth}
\centering
\begin{tabular}{llccccccccccccc }
\toprule
Precision &  block-size (W|A)  & OPT-6.7b & OPT-13b & OPT-30b & OPT-66b & BLM-1.7b & BLM-3b & BLM-7.1b & BLM-176b \\\midrule
W4A16   &   128 | NA     & 12.10     & 11.28   & 10.74   & 10.44   & 20.92    & 17.90   & 15.17    & 10.94     \\\midrule
\multirow{3}{*}{W4A8}    & Case-1: per-row | per-row & 13.17    & 13.07   & 14.65   & 16.32   & 21.43    & 18.39  & 15.58     & 11.19    \\
     & Case-2:  per-row  | 128  & 12.29    & 11.45   & 10.80    & 10.61   & 21.59    & 18.31  & 15.52    & 11.03    \\
     & Case-3: 128 | 128   & 12.04    & 11.31   & 10.75   & 10.45   & 21.27    & 17.86  & 15.19    & 10.96   \\
\bottomrule
\end{tabular}
\end{adjustbox}
\end{table}

\begin{wraptable}{r}{8cm}
\vspace{-0.15cm}
\caption{
\bloom-176B with different quantization block sizes on activation. 
Here weight is asymmetrically quantized with block size 128.
See more in~\tref{tab:bloom-176-different-blocks-full}.
}\centering
\label{tab:bloom-176-different-blocks}
\begin{adjustbox}{width=1.0\linewidth}
\centering
\begin{tabular}{lcccccccccccccc }
\toprule
A8 Block Size  & 1024   &512    &256 &128 &32 \\
\midrule
\ppl &10.98 &10.97 &10.95 &10.95  &10.95\\
\bottomrule
\end{tabular}
\end{adjustbox}
\vspace{-0.5cm}
\end{wraptable}

\textbf{Activation Quantization (W4A8).} To comprehend the benefits of fine-grained quantization on activation, we analyze the quantization between per-row and a block size of 128, with INT4 weight, as highlighted in \tref{tab:opt-4bit8bit-blocksize}. For models of considerable size, specifically those equal to or exceeding 1B, the application of such fine-grained activation quantization (Case-1) results in a substantial reduction in quantization error compared to per-row activation (Case-2). By implementing fine-grained activation quantization with weight quantization (Case-3), we are able to almost restore the performance to the level of their W4A16 counterparts. 

Furthermore, we detail the impacts of varying activation quantization block sizes in \tref{tab:bloom-176-different-blocks} on \bloom-176B, with INT4 weight. A trend of superior accuracy is observed with smaller block sizes in contrast to larger ones. However, the enhancement in performance reaches a saturation point when the size smaller or equal to 256, which corresponds to the range of values INT8 can represent. Despite INT8's capability to signify 256 distinct values, activation quantization errors persist due to the application of uniform quantization.

\begin{tcolorbox}
\textbf{Finding 3 on FGQ. }   \textbf{(1)} Larger models ($\geq$10B) are capable of attaining \classone error for 4-bit quantization. These models can leverage low-precision quantization as the model size with INT4 is similar to an INT8 model that is half its size, with improved accuracy. On the other hand, smaller models ($\leq$10B) typically reach only \classtwo or \classthree error levels.
    \textbf{(2)}  For larger models (>10B), the difference between fine-grained weight-and-activation quantization and fine-grained weight-only quantization is insignificant. 
    \textbf{(3)} The advantage of fine-grained activation quantization fades for larger models when the block size reaches 256.
\end{tcolorbox}
  \vspace{-0.3cm}
\section{Proposed Method to Further Push the Limit of Post-training Quantization}
\label{sec:design}
Building on the investigation and conclusions drawn from previous sections, it has become apparent that there is still a need for an advanced methodology to further refine the existing methods, with the objective of fully realizing the original fp16 \ppl quality. In this section, we introduce a simple yet effective method called \textbf{LoRC} (Low Rank Compensation) to optimize the current existing quantization error and further bridge the gap between the quality of the original model and its quantized counterparts.

LoRC is inspired by the employment of low-rank matrix factorization on the quantization error matrix $E:=W- \hat{W}$, where $W$ represents the original weight and $\hat{W}$ is the quantized weight. LoRC approximates the error $E$ with $\hat{E}=\hat{U}\hat{V}$ by using two low-rank matrices $\hat{U}$ and $\hat{V}$. This results in a more accurate approximation of the original weight matrix $W$ by $\hat{W}_{\text{lorc}} = \hat{W} + \hat{E}$, thereby reducing quantization errors: $\|W-\hat{W}\|\geq \|W-\hat{W}_{\text{lorc}}\|$. LoRC consists of two  steps:

\textbf{Step I:} Implement Singular Value Decomposition (SVD) on the error matrix $E = U \Sigma V$, where $U \in\mathbb{R}^{d_\text{in}\times d_\text{in}}$ and $V \in \mathbb{R}^{d_\text{out} \times d_\text{out}}$ are unitary matrices, and $\Sigma \in\mathbb{R}^{d_\text{in}\times d_\text{out}}$ is a diagonal matrix with its diagonal elements ordered in a descending manner.

\textbf{Step II:} We formulate the matrix $\hat{E} = \hat{U} \hat{V}$ where $\hat{U}= U_m(\Sigma_m)^{\frac{1}{2}}$ and $\hat{V}= (\Sigma_m)^{\frac{1}{2}} V_m$. Here, $U_m = U_{:, 1:m} \in\mathbb{R}^{d_\text{in}\times m}$, $V_m = V_{1:m, :} \in\mathbb{R}^{ m \times d_\text{out}}$, and $\Sigma_m = \Sigma_{1:m, 1:m} \in\mathbb{R}^{ m \times m}$.

The objective of LoRC is to achieve a good approximation of the error matrix $E$ using low-rank matrices, with minimal impact on the increase in model size. For instance, consider the standard transformer models~\cite{vaswani2017attention}, where each layer is comprised of a multi-headed attention (MHA) module and a multi-linear perception (MLP) module. Let $h$ represent the hidden dimension and $l$ the number of layers. The total number of parameters is $12lh^2$ as each layer contains $4h^2$ for MHA (for key, query, value, and projection matrices), and $8h^2$ for MLP (two matrices of sizes $h\times 4h$ and $4h\times h$). With the addition of low-rank LoRC to the six matrices in each layer, the total number of parameters for $l$ layers would amount to $18hml$.\footnote{In the MHA module, LoRC contributes $2hm$ to each of key, query, value, and the projection matrices. In the MLP module, LoRC contributes $8hm$ and $2hm$ respectively to the matrices of dimensions $h\times 4h$ and $4h\times h$.} Consequently, the ratio of parameters added to the existing model is $3m/2h$. It's important to note that the low-rank dimension $m$ can be as small as $4$ or $8$ (which we will discuss in detail in a later section) while the standard hidden dimension $h\ge 768$, making the number $3m/2h\leq 0.016$. 


Significantly, LoRC can be viewed as a supplementary feature to existing quantization methodologies such as RTN, GPTQ, and ZeroQuant-Local/Global, and can be seamlessly integrated with FGQ. We have conducted experiments to evaluate the performance of LoRC on both \opt and \bloom, applying 4-bit, 3-bit, and 2-bit weights by setting the activation to FP16.\footnote{For INT8 Activation, please see \tref{tab:LORC-int8}, the observation for FP16 holds similarly for INT8 Activation.} Based on the discoveries in the preceding sections, we utilize the GPTQ quantization strategy. To gain a comprehensive understanding of LoRC, we include the results with and without the application of FGQ. The datasets and hyperparameters are consistent with those detailed in earlier sections.


\begin{table}
\caption{
 W\#\asym-A16 quantization with \# being 4-bit, 3-bit and 2-bit on \opt and \bloom (BLM). 
}\centering
\label{tab:lorc-4bit-blocksize}
\begin{adjustbox}{width=1.01\linewidth}
\centering
\begin{tabular}{lc|ccccc|ccccccc }
\toprule
 \multirow{2}{*}{Bits}            & \multirow{2}{*}{\small \textbf{LoRC}}       & \multicolumn{5}{c|}{Coarse-grained weight quantization (per-row block-size)}     & \multicolumn{5}{c}{Fine-grained quantization on weight (256 block-size )}                \\
                  &   &  \small OPT-6.7b & \small OPT-13b &  \small OPT-30b &  \small OPT-66b & \small BLM-176b & \small OPT-6.7b & \small OPT-13b &  \small OPT-30b &  \small OPT-66b & \small BLM-176b \\\midrule
    \multicolumn{2}{c|}{W8A16}                      & 11.90     & 11.22   & 10.70    & 10.33   & 10.90       & 11.90     & 11.22   & 10.70    & 10.33   & 10.90       \\\midrule
\multirow{2}{*}{W4A16} & \xmark     & 12.28    & 11.42   & 10.78   & 10.78   & 11.02      & 12.05    & 11.28   & 10.74   & 10.50    & 10.95      \\
                       & \cmark    & 12.10    & 11.36   & 10.76   & 10.34   & 10.98      & 11.99    & 11.29   & 10.70    & 10.29   & 10.93      \\ \midrule
\multirow{2}{*}{W3A16} & \xmark     & 14.18    & 12.43   & 11.28   & 17.77   & 49.46      & 12.79    & 11.63   & 10.9    & 11.34   & 11.13      \\
                       & \cmark    & 13.00     & 11.90    & 11.14   & 10.63   & 11.30       & 12.40    & 11.57   & 10.83   & 10.42   & 11.08      \\\midrule
\multirow{2}{*}{W2A16} & \xmark    & 120.56   & 40.17   & 25.74   & 225.45  & Explode    & 23.13    & 15.55   & 12.68   & 308.49  & 12.64      \\
                       & \cmark    & 24.17    & 18.53   & 14.39   & 13.01   & 14.15      & 16.27    & 14.30   & 12.37   & 11.54   & 12.21    \\
\bottomrule
\end{tabular}
\end{adjustbox}
\end{table}

\textbf{Evaluation Results.} The findings are showcased in \tref{tab:lorc-4bit-blocksize}, split into two sections: coarse-grained weight quantization (per-row) and fine-grained quantization (block-size 256). Notably, we observe that the two low-rank matrices, $\hat{U}$ and $\hat{V}$, can be quantized to 8-bit without any performance discrepancy (\tref{tab:lorc-int8}). Thus, the two low-rank matrices for LoRC in \tref{tab:lorc-4bit-blocksize} are INT8 with a low-rank dimension of $m=8$. 

\begin{wraptable}{r}{8cm}
\caption{
\small Results of W4\asym A16 quantization with LoRC approximating $\hat{E}=\hat{U}\hat{V}$ on OPT model family.  $\hat{U}$ and $\hat{V}$ can be represented with FP16 or INT8, of which the performance are represented below. There is hardly any difference between FP16 and INT8.  
}\centering
\label{tab:lorc-int8}
\begin{adjustbox}{width=1.01\linewidth}
\centering
\begin{tabular}{l|cccc|ccccccccc }
\toprule
LoRC          & \multicolumn{4}{c|}{Coarse-grained weight quantization} & \multicolumn{3}{c}{Fain-grained weight Quantization} \\
$\hat{U},\hat{V}$           & 6.7b           & 13b          & 30b          & 66b          & 6.7b                & 13b                & 30b                \\\midrule
 FP16 & 12.08              & 11.35            & 10.76            & 10.31            & 11.993                  & 11.290                 & 10.703                 \\
INT8 & 12.10              & 11.36            & 10.76            & 10.34            & 11.987                  & 11.290                 & 10.700   \\
\bottomrule
\end{tabular}
\end{adjustbox}
\end{wraptable}

Several key observations can be made. Firstly, LoRC consistently boosts performance across all bit sizes and block sizes, as indicated by the lower perplexity scores when LoRC is activated. Secondly, the enhancement brought about by LoRC becomes more substantial as the bit size diminishes, especially noticeable for W2A16, which displays a markedly greater impact compared to W4A16 and W3A16 in most scenarios. Lastly, the combination of fine-grained quantization with LoRC yields the most impressive results, underscoring the efficacy of LoRC when integrated with FGQ.
Overall, the results emphasize the benefits of using LoRC for enhanced performance in weight quantization and its compatibility with FGQ. Notably, recovering the last 0.05-0.1 perplexity can be challenging, but with LoRC, we are able to nearly recover the original model quality for INT4 quantization.
\begin{table}[t]
\vspace{-0.05cm}
\begin{minipage}[c]{0.4\textwidth}
     \centering
\vspace{-0.55cm}
\caption{ \small W4A16 quantization with LoRC by varying the low-rank dimension $m$. }\centering
\label{tab:lorc-dim}
\begin{adjustbox}{width=0.99\linewidth}
\centering
\begin{tabular}{l|cccccccc }
\toprule
LoRC-dim $m$ & OPT-1.3b  & OPT-6.7b  &  OPT-30b   \\\midrule 
$m=0$ basline          & 15.95 & 12.06 & 10.73 \\\midrule 
$m=1$          & 15.93 & 12.01 & 10.73 \\
 $m=4$            & 15.73 & 12.00 & 10.72 \\
$m=8$         & 15.76 & 11.99 & 10.70 \\
$m=16$         & 15.74 & 12.00 & 10.69 \\
$m=32$          & 15.71 & 12.01 & 10.69\\
\bottomrule
\end{tabular}
\end{adjustbox}
%
\end{minipage}
\begin{minipage}[c]{0.59\textwidth}
\centering
\includegraphics[width=0.85\textwidth]{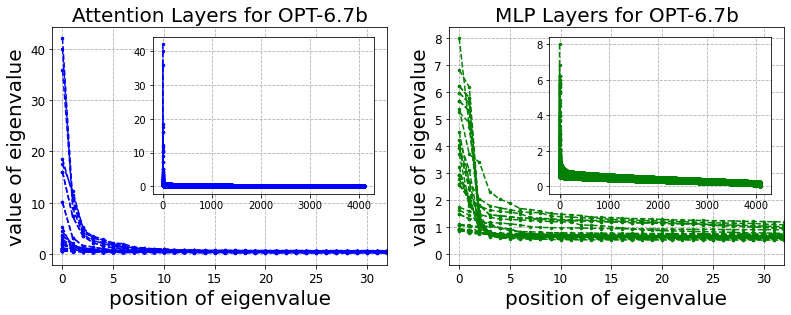}
 \captionof{figure}{ \small Eigenvalues of the Error matrix $E$ for W4A16}\label{fig:Eigenvalues}
\end{minipage}
\end{table}

\textbf{Ablation Study on the Low Rank Dimension $m$.} An essential aspect of the LoRC method is on the optimal low-rank dimension, denoted as $m$, explained in \textbf{Step II}. To explore this, we varied $m$ in the range of 1, 4, 8, 16, and 32 for OPT-1.3b/6.7b/30b models, and applied W4A16 GPTQ quantization. The outcomes are depicted in Table \ref{tab:lorc-dim}, indicating that the enhancements achieved through LoRC begin to plateau as the dimension $m$ surpasses 4. The most optimal performance for OPT-6.7b is realized when $m=8$.

This observation may seem counterintuitive initially, as one might anticipate that larger LoRC dimensions would yield more significant improvements. To gain a more comprehensive understanding, we conducted an analysis of the eigenvalues of the actual error matrix $E=W-\hat{W}$ for each matrix. By randomly selecting 20 matrices from MHA and MLP layers, we plotted the eigenvalues of $E$ as a curve, depicted in \fref{fig:Eigenvalues}. The two plots reveal a rapid flattening of eigenvalues after index 8, which elucidates why increasing the LoRC dimension does not considerably enhance performance. Hence, a sensible dimension for $\hat{U}$ and $\hat{V}$ in the LoRC methodology could be 8.\footnote{Please note that this observation is only true for PTQ. If one uses quantize-aware training (QAT) and let $\hat{U}$ and $\hat{V}$ updated during QAT, we arrive at contrasting conclusions. For more details, please refer to \appref{sec:qat-lorc}.}



\section{Discussion}
\label{sec:conclusions}
\paragraph{Conclusion.}
In this work, we provide a comprehensive study of post-training quantization (\ptq) on large language models with different \ptq methods (e.g., \rtn, \gptq, ZeroQuant), and with different quantization coverage (weight-only and weight-and-activation quantization), etc. 
We find that \ptq methods are critical to improving the quantized model quality, and that fine-grained quantization (FGQ) can bring acceptable accuracy and model size trade-off. Finally, we introduced an optimization technique called Low Rank Compensation (LoRC), which works synergistically with PTQ and FGQ, playing a crucial role in enhancing full model quality recovery with a minimal increase in model size.

\paragraph{Limitation.} Despite quantizing over 10,000 experiments, our study was constrained by our computing resources. This restriction made us choose between diversifying the model sizes and varying the tasks. We strategically limited our datasets to WikiText, PTB, and C4 to concentrate on a broad range of quantization methods. Consequently, our general findings are more robust concerning the two model families and three datasets examined in this paper. However, caution should be exercised when generalizing these findings to tasks that are dissimilar to those covered in this study.
\paragraph{Future Opportunity.}
Throughout the paper, we see several unresolved problems from current quantization schemes and/or algorithms, and we find potential directions for \llm compression: (1) Although we use fine-grained quantization schemes in the paper, the real implementation is missing. Moreover, how to efficiently implement odd bit precision is  challenging. 
    \cite{frantar2022gptq} demonstrated that 3-bit can achieve better throughput in the generation phase by packing all 3-bit numbers in continuous memory space. 
    However, this method is sub-optimal as the dequantization step needs to connect bits from different bytes.
    One possible way to implement odd bits, e.g., 5 bits, is to use two integer matrices with INT4 and INT1.
    During the dequantization stage, we couple the two matrices together. 
    (2) How to combine \ptq with other lightweight compression techniques, e.g., post-training pruning~\cite{kwon2022fast,frantar2023massive}, is an interesting direction to further reduce the memory consumption and compute cost.


{
\bibliographystyle{plain}
\bibliography{ref.bib}
}

\clearpage
\onecolumn
\appendix
\section{Background of Quantization}
\label{sec:background_of_quantization}
Quantization maps floating point (e.g., FP16/FP32) numbers to integer numbers (e.g., INT4/INT8) so that lower memory usage (weight quantization) and faster integer arithmetic (weight-and-activation quantization) can be achieved compared to the floating point format. 
In this work, we are focusing on uniform quantization, i.e., 
\begin{equation}
\small
\label{eq:quantization_formula}
Q(x) = \text{INT}\big({(x-Z)}/{S}\big)-Z,
\end{equation}
where $Q$ is the quantization function, $x$ is a floating point input vector/tensor, $S$ is a real valued scaling factor, and $Z$ is an integer zero point. 
Based on different settings, the quantization method can be viewed as (1) symmetric vs. asymmetric quantization ($Z=0$ or not), (2) fine-grained vs. coarse-grained quantization (how to partition the input x and get its associated scaling factor, e.g., matrix wise or row wise). 
See~\cite{gholami2021survey} for more details. 

Throughout this work, we focus on post-training quantization (\ptq), i.e., no or minimal training effort is applied after quantization, for which large accuracy degradation usually exhibits for coarse-grained quantization (per matrix/tensor) due to their large quantization error.
As such, we focus on fine-grained quantization.
Particularly, we use the per-row quantization (one row of the weight matrix or one token for the activation) from~\cite{yao2022zeroquant} as our coarsest-grained quantization method, 
and we use block-k quantization (for every k elements, they have their own scaling factor and/or zero point) as our finer-grained quantization scheme. 

\counterwithin{figure}{section}
\counterwithin{table}{section}

\section{Detailed Setting Used in~\sref{sec:evaluation_of_existing_methods}}
\label{sec:hyperparameter_used_in_exisiting_method_evaluation}
Same as~\cite{frantar2022gptq}, for all methods, we use C4 dataset to randomly select 128 sentences for training and each of them has 2048 tokens.

For \gptq, we check its main hyperparameter, i.e., the dampening factor, and find out the method is not sensitive to it.
As such, we use the hyparameter suggested by the author for all of our experiments. 
For \zqglobal and \zqlocal, as mentioned the in main text, the hyperparameters suggested by the original work~\cite{yao2022zeroquant} is suboptimal. 
We find that a linear decay learning rate schedule is very helpful in our initial test.
As such, we add this as our default setting.
Meanwhile, we extensively test a wide range (1e-3 to 5e-8) of learning rate for different models until we find the best learning rate (i.e., larger or smaller learning rate leads to worse accuracy performance).We employed the Adam optimizer and set the default batch size to 1 for our experiments. 

We conducted tests to assess whether changes in random seeds would introduce substantial variations in the outcomes. As per the findings detailed in Table \tref{tab:mean}, the modifications in random seeds resulted in only minimal effects on the final quality of the models. This effect was particularly negligible in the context of larger models, such as OPT-30b, where the standard deviation was only 0.01. Therefore, in consideration of these results, we elected to standardize the random seed for the subsequent experiments presented in this paper, setting it uniformly at 123 or 0. The code will be made publicly available to facilitate reproducibility of our results. 

For all three methods, we run them on a single GPU (either V100-32GB or A100-80GB). 
For the largest model tested in the paper, i.e., \bloom-176B, the cost of all methods is lower than one GPU-day on A100-80G.

\begin{table}[H]
\caption{The table on the left illustrates the outcomes of each task, evaluated using three different random seeds. On the right, we present a table detailing the mean and standard deviation of the Task-mean values (which can be found in the final column of the left table) over the three random seeds, accompanied by additional quantization results. The quantization methodologies employed in this context are based on the \gptq algorithm.
}\centering
\label{tab:mean}
\begin{adjustbox}{width=0.99\linewidth}
\centering
\begin{tabular}{lcccccccccccccc }
\toprule
Precision              & Random Seed & WikiText  & PTB  & C4    & Task-mean \\\midrule
\multirow{2}{*}{OPT-13b} &123  & 10.31 & 12.62 & 11.35 & 11.43     \\
\multirow{2}{*}{W4A16}&234  & 10.25 & 12.57 & 11.35 & 11.39     \\
&456  & 10.37 & 12.61 & 11.36 & 11.44     \\\midrule
\multirow{2}{*}{OPT-30b} &123  & 9.56  & 11.95 & 10.79 & 10.77     \\
\multirow{2}{*}{W4A16}&234  & 9.6   & 11.95 & 10.79 & 10.78     \\
&456  & 9.52  & 11.97 & 10.79 & 10.76  \\
\bottomrule
\end{tabular}
\begin{tabular}{lcccccccccccccc }
\toprule

Precision              & Items & OPT-1.3b & OPT-13b & OPT-30b \\\midrule
\multirow{2}{*}{W4A16} & mean over three random seeds  & 16.39    & 11.42   & 10.77   \\
                       & standard deviation   & 0.019    & 0.027   & 0.010   \\\midrule
\multirow{2}{*}{W4A8}  & mean over three random seeds  & 16.76    & 17.16   & 21.64   \\
                       & standard deviation    & 0.048    & 0.048   & 1.277  \\
\bottomrule
\end{tabular}

\end{adjustbox}
\end{table}

\section{Best \ptq Methods with Per-row Quantization}
Table~\ref{tab:opt-quantization-method} and~\ref{tab:bloom-quantization-method} summarize the best \ptq methods with per-row optimization.

\begin{table}[t]
\caption{
Best optimization method of \opt family in~\sref{sec:evaluation_of_existing_methods}.
}\centering
\label{tab:opt-quantization-method}
\begin{adjustbox}{width=0.9\linewidth}
\centering
\begin{tabular}{lcccccccccccccc }
\toprule
Precision     & 125m	& 350m 	& 1.3b	& 2.7b	& 6.7b	& 13b &30b	& 66b \\
\midrule
Weight Only (INT4) &\zqglobal &\zqglobal &\gptq &\gptq &\gptq &\gptq &\gptq &\gptq\\
\midrule
Weight \& Activation (W4A8) &\zqglobal &\zqglobal &\zqglobal &\gptq &\zqglobal &\zqglobal &\zqglobal &\zqlocal\\
\bottomrule
\end{tabular}
\end{adjustbox}
\end{table}

\begin{table}[t]
\caption{
Best optimization method of \bloom family in~\sref{sec:evaluation_of_existing_methods}.
}\centering
\label{tab:bloom-quantization-method}
\begin{adjustbox}{width=0.9\linewidth}
\centering
\begin{tabular}{lcccccccccccccc}
\toprule
Precision     & 560m   &1.1b   & 1.7b  & 3b & 7.1b & 176b \\
\midrule
Weight Only (INT4) &\gptq &\zqglobal &\zqglobal &\zqglobal/\gptq &\gptq &\gptq\\
\midrule
Weight \& Activation (W4A8) &\zqglobal &\zqglobal &\zqglobal &\zqglobal &\zqglobal &\zqlocal \\
\bottomrule
\end{tabular}
\end{adjustbox}
\end{table}

\section{Quantization-aware training with LoRC}\label{sec:qat-lorc}
In order to better understand our proposed algorithm, \lorc, particularly in relation to the dimensions of low-rank matrices, we applied quantize-aware training alongside knowledge distillation. This approach builds upon the methodology of row-wise weight quantization and token-wise quantization. For the optimization process, we employed the Adam optimizer, setting the learning rate at 1e-4 and a dropout rate of 0.05. These settings were identified as the most effective in our context (additional details can be found in \cite{wu2023understanding}). We performed fine-tuning on the WikiText dataset using pre-trained GPT2 models with 125M and 350M parameters, which were obtained from Hugging Face as our initial models. \footnote{\url{https://huggingface.co/gpt2}}

\begin{figure}[H]
\centering
\includegraphics[width=0.35\textwidth]{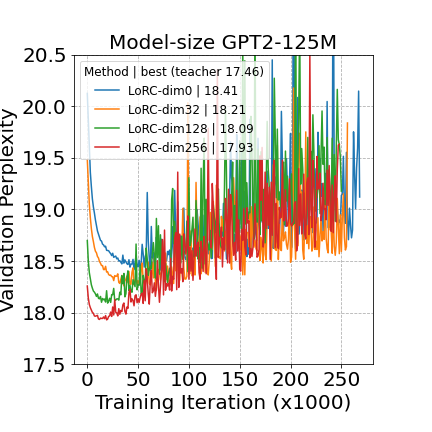}
\includegraphics[width=0.35\textwidth]{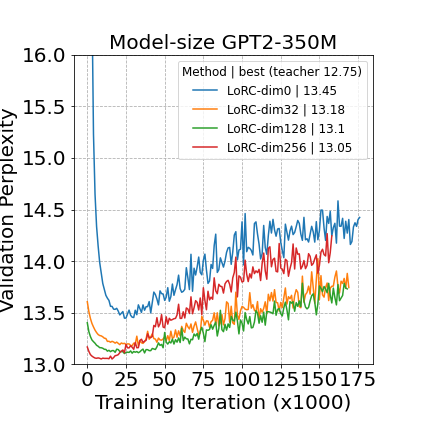}
 \captionof{figure}{The graph on the left represents the results for a smaller model size (GPT2-125M), while the one on the right corresponds to the GPT2-350M model. The dimension (refer to the legend) in the LoRC algorithm, which is represented by different color curves, plays a pivotal role in approximating the original quality of the fp16 model.}\label{fig:lorc-despription}
\end{figure}

The results are illustrated in Figure \fref{fig:lorc-despription}. As observed, the quantized models tend to overfit swiftly. However, implementing higher dropout values, such as 0.1, does not result in a significantly improved performance with regards to the best perplexity over the entire training duration. Now when examining the best perplexity associated with each dimension of \lorc (also indicated in the figure's legend), it becomes evident that the larger the dimension, the better the W4A8 models perform. This suggests that augmenting the dimension of \lorc can enhance the model quality for QAT, a finding that deviates from the trends observed in PTQ.

\section{Tables and Figures}
\label{sec:full_tables_results}
We put the full results of our evaluations in this section.
\begin{table}[t]
\caption{
\opt ppl on wikitext/ptb/c4 (full results of~\tref{tab:ptq_challenge_opt_average_in_maintext}).
}\centering
\label{tab:ptq_challenge_opt_full_in_appendix}
\begin{adjustbox}{width=0.9\linewidth}
\centering
\begin{tabular}{lcccccccccccccc }
\toprule
Precision     & 125m	& 350m 	& 1.3b	& 2.7b	& 6.7b	& 13b &30b	& 66b \\
\midrule
W16-A16 &27.65/32.55/24.61 &22.00/26.08/20.71 &14.62/16.97/14.72 &12.47/15.11/13.17 &10.86/13.09/11.74 &10.13/12.34/11.20 &9.56/11.84/10.69 &9.34/11.36/10.28\\
\midrule
W8A8\sym-A16 &27.64/32.53/24.65 &22.06/26.10/20.72 &14.63/16.98/14.73 &12.48/15.13/13.17 &10.85/13.11/11.75 &10.12/12.34/11.20 &9.55/11.85/10.70 &9.34/11.36/10.29 \\
W8\asym-A16  &27.71/32.58/24.64 &22.04/26.12/20.73 &14.67/16.99/14.73 &12.50/15.14/13.17 &10.86/13.11/11.75 &10.11/12.34/11.20 &9.55/11.84/10.69 &9.35/11.36/10.29\\
W4\sym-A16 &45.89/53.68/36.68 &25.95/31.11/23.94 &19.85/23.61/18.90 &22.86/30.01/22.29 &12.41/17.05/13.62 &11.06/14.90/12.23 &10.18/13.26/11.86 &57.73/134.91/98.51 \\
W4\asym-A16  &36.71/44.76/30.92 &25.51/30.90/23.86 &19.38/21.95/17.93 &17.92/22.48/18.32 &11.91/15.39/13.01 &10.67/13.53/12.07 &10.10/13.13/11.33 &20.24/48.45/25.86 \\
\midrule
W16-A8\sym &27.96/32.57/24.69 &22.06/26.42/20.95 &15.21/18.18/15.81 &12.98/16.01/13.89 &20.99/25.94/31.18 &3341.50/2618.38/3554.59 &1681.48/2221.62/2241.53 &2696.91/2647.41/2569.94 \\
W16-A8\asym &27.84/32.60/24.66 &22.04/26.22/20.81 &15.14/17.65/15.39 &12.51/15.38/13.38 &11.24/14.17/12.45 &11.83/18.87/15.39 &14.08/31.54/25.09 &442.66/524.57/716.83  \\
\bottomrule
\end{tabular}
\end{adjustbox}
\end{table}

\begin{table}[t]
\caption{
\bloom ppl on wikitext/ptb/c4 (full results of~\tref{tab:ptq_challenge_opt_average_in_maintext}).
}\centering
\label{tab:ptq_challenge_bloom_full_in_appendix}
\begin{adjustbox}{width=0.9\linewidth}
\centering
\begin{tabular}{lcccccccccccccc }
\toprule
Precision     & 560m   &1.1b   & 1.7b  & 3b & 7.1b & 176b \\
\midrule
W16-A16  &22.43/41.25/24.38 &17.69/46.98/20.29 &15.39/27.93/17.97 &13.48/23.12/16.14 &11.37/19.40/14.13  &8.11/13.62/10.97 \\
\midrule
W8\sym-A16 &22.44/41.28/24.39 &17.70/47.01/20.29 &15.40/27.91/17.98 &13.49/23.14/16.14 &11.37/19.40/14.13 &8.11/13.63/10.98 \\
W8\asym-A16 &22.43/41.24/24.40 &17.69/47.00/20.29 &15.40/27.96/17.97 &13.48/23.14/16.14 &11.37/19.40/14.13 &8.10/13.62/10.98 \\
W4\sym-A16 &26.49/49.73/27.98 &20.27/56.64/22.81 &17.47/32.20/19.88 &14.96/25.59/17.51 &12.38/21.36/15.06 &8.40/14.15/11.30 \\
W4\asym-A16 &25.31/46.79/27.10 &23.90/68.31/25.99 &16.93/31.02/19.47 &14.65/25.12/17.26 &12.06/20.83/14.83 &8.34/14.03/11.23 \\
\midrule
W16-A8\sym &22.50/41.58/24.46 &17.78/47.28/20.38 &15.57/28.36/18.13 &13.57/23.38/16.25 &11.58/19.92/14.35 &8.75/14.94/12.61 \\
W16-A8\asym &22.45/41.37/24.42 &17.71/47.05/20.32 &15.45/28.09/18.02 &13.52/23.24/16.19 &11.47/19.71/14.25 &8.41/14.52/11.93 \\
\bottomrule
\end{tabular}
\end{adjustbox}
\end{table}

\begin{table}[t]
\caption{
\opt ppl on wikitext/opt/c4 with W4\asym-A16 (full table of~\tref{tab:weight_only_quantization_opt_existing_method_average_in_main_text}).
See~\tref{tab:weight_only_quantization_opt_existing_method_full_zqlocal} for all learning rate results of \zqlocal and~\tref{tab:weight_only_quantization_opt_existing_method_full_zqglobal} of \zqglobal.
}\centering
\label{tab:weight_only_quantization_opt_existing_method_full_in_appendix}
\begin{adjustbox}{width=0.9\linewidth}
\centering
\begin{tabular}{lcccccccccccccc }
\toprule
Precision     & 125m	& 350m 	& 1.3b	& 2.7b	& 6.7b	& 13b &30b	& 66b \\
\midrule
\rtn &36.71/44.76/30.92 &25.51/30.90/23.86 &19.38/21.95/17.93 &17.92/22.48/18.32 &11.91/15.39/13.01 &10.67/13.53/12.07 &10.10/13.13/11.33 &20.24/48.45/25.86 \\
\gptq &32.52/40.25/27.78 &23.50/29.14/22.41 &15.52/18.16/15.56 &13.02/15.84/13.73 &11.16/13.59/12.08 &10.29/12.61/11.35 &9.61/11.95/10.79 &9.54/11.67/10.52\\
\zqlocalstar &33.05/39.34/28.11 &24.40/29.22/22.82 &15.81/18.66/15.76 &13.22/16.19/13.96 &11.32/13.79/12.26 &10.42/12.90/11.60 &9.97/12.32/11.03 &9.91/11.87/10.59 \\
\zqglobalstar &31.44/36.66/27.21 &23.32/28.05/21.98 &15.46/18.31/15.67 &13.03/16.04/13.83 &11.30/13.69/12.17 &10.38/12.85/11.62 &9.90/12.24/10.99 &9.62/11.81/10.61 \\
\bottomrule
\end{tabular}
\end{adjustbox}
\end{table}

\begin{table}[t]
\caption{
\opt ppl on wikitext/opt/c4 with W4\asym-A16 and \zqlocal. 
}\centering
\label{tab:weight_only_quantization_opt_existing_method_full_zqlocal}
\begin{adjustbox}{width=0.9\linewidth}
\centering
\begin{tabular}{lcccccccccccccc }
\toprule
LR (W4\asym-A16)     & 125m	& 350m 	& 1.3b	& 2.7b	& 6.7b	& 13b &30b	& 66b \\
\midrule
0.001 &33.67/39.45/29.11 &26.33/31.94/24.49 &16.27/19.91/16.46 &14.34/17.76/14.93 &11.87/15.04/13.06 &13.68/18.89/14.46 &171.35/151.55/46.14 &814.22/601.74/308.53  \\
0.0005 &32.76/39.51/28.64 &25.88/30.95/23.96 &16.29/19.82/16.27 &14.16/17.65/14.79 &11.92/15.23/12.95 &10.93/13.82/12.03 &10.23/13.46/11.44 &10.10/12.27/10.81  \\
0.0001 &33.86/40.01/28.29 &24.64/30.26/23.33 &16.07/19.25/15.93 &14.36/17.38/14.41 &11.85/14.64/12.74 &10.93/13.48/11.88 &10.18/12.67/11.13 &10.12/12.01/10.67  \\
5e-05 &33.05/39.34/28.11 &25.42/29.65/23.22 &15.79/19.16/15.88 &13.70/16.80/14.16 &11.71/14.32/12.41 &10.75/13.38/11.77 &9.95/12.54/11.09 &10.02/11.89/10.64  \\
1e-05 &33.78/40.41/28.84 &24.40/29.22/22.82 &15.81/18.66/15.76 &13.55/16.46/13.96 &11.32/13.79/12.26 &10.54/13.05/11.61 &9.98/12.22/10.99 &9.91/11.87/10.59  \\
5e-06 &34.47/41.04/29.02 &24.50/29.27/23.00 &16.01/18.73/15.91 &13.22/16.19/13.96 &11.33/13.86/12.29 &10.42/12.90/11.60 &9.86/12.33/10.97 &9.97/11.86/10.60  \\
1e-06 &35.88/43.69/30.35 &24.54/29.87/23.17 &16.77/19.45/16.47 &13.60/17.02/14.46 &11.41/14.10/12.41 &10.53/13.01/11.70 &9.97/12.33/11.04 &10.01/11.93/10.66  \\
\bottomrule
\end{tabular}
\end{adjustbox}
\end{table}

\begin{table}[t]
\caption{
\opt ppl on wikitext/opt/c4 with W4\asym-A16 and \zqglobal. 
NaN here means the PPL is larger than 1e6.
}\centering
\label{tab:weight_only_quantization_opt_existing_method_full_zqglobal}
\begin{adjustbox}{width=0.9\linewidth}
\centering
\begin{tabular}{lcccccccccccccc }
\toprule
LR (W4\asym-A16)     & 125m	& 350m 	& 1.3b	& 2.7b	& 6.7b	& 13b &30b	& 66b \\
\midrule
0.001 &4057.13/2718.91/1247.78 &5071.35/5229.93/687.35 &12105.25/10154.73/7893.43 &18965.76/17112.60/16316.31 &60014.66/56041.86/78085.84 &232421.09/98305.32/119762.73 &93917.09/70170.34/51124.06 & NaN  \\
0.0005 &31.94/38.61/27.17 &27.11/33.91/24.07 &10900.84/8322.65/8425.10 &14412.30/8676.76/10154.55 &18527.46/13530.12/13029.95 &109006.53/62584.41/125349.50 &303235.75/230599.62/430480.03 &36439.32/30554.19/33756.93  \\
0.0001 &31.44/36.66/27.21 &24.08/29.08/22.27 &15.91/20.08/16.35 &118.38/53.47/54.08 &7604.92/5339.10/5161.49 &12638.86/7639.95/8243.63 &16276.68/9890.26/6176.27 &8367.31/4728.13/5533.59  \\
5e-05 &31.97/36.93/27.12 &23.55/28.06/22.02 &15.82/18.65/15.65 &13.40/16.44/13.97 &26.54/25.67/17.60 &909.99/316.82/370.84 &6238.21/3291.04/3743.01 &9296.98/6687.44/5363.29  \\
1e-05 &32.31/37.93/27.38 &23.32/28.05/21.98 &15.60/18.42/15.64 &13.09/16.05/13.78 &11.41/13.82/12.20 &10.80/13.16/11.66 &10.06/12.44/11.07 &9.73/12.09/10.98  \\
5e-06 &32.69/38.91/27.76 &23.26/28.33/22.05 &15.46/18.31/15.67 &13.03/16.04/13.83 &11.30/13.69/12.17 &10.50/12.89/11.58 &9.95/12.28/11.01 &9.62/11.81/10.61  \\
1e-06 &34.63/41.75/29.43 &23.82/28.96/22.48 &16.12/19.46/16.27 &13.03/16.27/14.04 &11.29/13.88/12.27 &10.38/12.85/11.62 &9.90/12.24/10.99 &9.58/12.17/10.78  \\
5e-07 & NaN & NaN & NaN & NaN & NaN &10.51/12.96/11.70 &9.89/12.41/11.04 &9.90/12.45/11.00  \\
1e-07 & NaN & NaN & NaN & NaN & NaN &10.63/13.29/11.89 &10.02/12.82/11.18 &11.03/13.91/11.73  \\
5e-08 & NaN & NaN & NaN & NaN & NaN &10.66/13.42/11.97 &10.05/13.00/11.24 &12.41/17.45/13.02  \\
\bottomrule
\end{tabular}
\end{adjustbox}
\end{table}

\begin{table}[t]
\caption{
\bloom ppl on wikitext/opt/c4 with W4\asym-A16 (full table of~\tref{tab:weight_only_quantization_opt_existing_method_average_in_main_text}).
See~\tref{tab:weight_only_quantization_opt_existing_method_full_zqlocal} for all learning rate results of \zqlocal and~\tref{tab:weight_only_quantization_opt_existing_method_full_zqglobal} of \zqglobal.}\centering
\label{tab:weight_only_quantization_bloom_existing_method_full_in_appendix}
\begin{adjustbox}{width=0.9\linewidth}
\centering
\begin{tabular}{lcccccccccccccc}
\toprule
Precision     & 560m   &1.1b   & 1.7b  & 3b & 7.1b & 176b \\
\midrule
\rtn &25.31/46.79/27.10 &23.90/68.31/25.99 &16.93/31.02/19.47 &14.65/25.12/17.26 &12.06/20.83/14.83 &8.34/14.03/11.23 \\
\gptq &23.90/43.76/25.59 &24.34/68.10/26.58 &16.36/29.58/18.79 &14.10/24.23/16.66 &11.80/20.23/14.47 &8.22/13.78/11.07 \\
\zqlocalstar &24.23/44.94/26.05 &19.22/52.36/21.59 &16.37/29.89/18.86 &14.23/24.41/16.86 &11.80/20.28/14.56 &8.27/13.91/11.16 \\
\zqglobalstar &23.84/44.17/25.60 &19.50/51.33/21.72 &16.19/29.28/18.66 &14.14/24.16/16.69 &11.77/20.27/14.52 &8.24/13.82/11.10 \\
\bottomrule
\end{tabular}
\end{adjustbox}
\end{table}

\begin{table}[t]
\caption{
\bloom ppl on wikitext/opt/c4 with W4\asym-A16 and \zqlocal. 
}\centering
\label{tab:weight_only_quantization_bloom_existing_method_full_zqlocal}
\begin{adjustbox}{width=0.9\linewidth}
\centering
\begin{tabular}{lcccccccccccccc}
\toprule
LR (W4\asym-A16)     & 560m   &1.1b   & 1.7b  & 3b & 7.1b & 176b \\
\midrule
0.001 &25.37/47.36/27.03 &19.89/53.86/22.11 &16.70/31.19/19.30 &14.45/25.28/17.16 &12.22/21.34/15.04 &8.82/15.77/11.98  \\
0.0005 &25.17/46.83/26.87 &19.57/53.66/21.92 &16.58/30.27/19.15 &14.43/25.47/17.07 &11.94/20.54/14.67 &8.35/14.01/11.20  \\
0.0001 &24.59/46.11/26.32 &19.22/52.36/21.59 &16.41/30.29/18.90 &14.35/24.81/16.87 &11.83/20.34/14.58 &8.28/13.92/11.14  \\
5e-05 &24.44/46.04/26.16 &23.28/65.68/25.42 &16.39/30.01/18.86 &14.34/24.43/16.83 &11.80/20.28/14.56 &8.27/13.93/11.15  \\
1e-05 &24.23/44.94/26.05 &23.45/66.29/25.52 &16.37/29.89/18.86 &14.23/24.41/16.86 &11.84/20.39/14.58 &8.27/13.91/11.16  \\
5e-06 &24.21/45.21/26.10 &23.26/65.72/25.42 &16.42/30.09/18.94 &14.25/24.55/16.87 &11.87/20.50/14.61 &8.29/13.98/11.16  \\
1e-06 &24.71/45.86/26.50 &23.45/66.28/25.56 &16.64/30.52/19.15 &14.46/24.76/17.04 &11.94/20.55/14.70 &8.29/13.97/11.18  \\
\bottomrule
\end{tabular}
\end{adjustbox}
\end{table}

\begin{table}[t]
\caption{
\bloom ppl on wikitext/opt/c4 with W4\asym-A16 and \zqglobal. 
}\centering
\label{tab:weight_only_quantization_bloom_existing_method_full_zqglobal}
\begin{adjustbox}{width=0.9\linewidth}
\centering
\begin{tabular}{lcccccccccccccc}
\toprule
LR (W4\asym-A16)     & 560m   &1.1b   & 1.7b  & 3b & 7.1b & 176b \\
\midrule
0.001 &6853935.00/30441738.00/3222857.25 &528072.88/828428.62/356031.97 &597410.50/973155.88/1280478.12 &878460.69/2175974.25/441401.94 &nan/nan/nan & NaN  \\
0.0005 &29671.52/1795030.88/4653.35 &28112.96/87515.64/1826.82 &141110.14/204295.86/40146.11 &265457.25/741326.38/99882.45 &944784.19/774538.25/395960.03 & NaN  \\
0.0001 &23.92/45.68/25.72 &19.34/52.78/21.63 &16.35/29.22/18.76 &14.27/24.46/16.80 &12.17/22.16/14.80 & NaN  \\
5e-05 &23.84/44.17/25.60 &19.50/51.33/21.72 &16.19/29.28/18.66 &14.14/24.16/16.69 &11.81/20.41/14.50 & NaN  \\
1e-05 &23.85/44.20/25.65 &22.64/56.79/23.41 &16.23/29.73/18.73 &14.14/24.31/16.74 &11.77/20.27/14.52 &8.24/13.82/11.10  \\
5e-06 &24.02/44.62/25.79 &23.46/63.27/24.88 &16.28/29.83/18.81 &14.19/24.38/16.80 &11.77/20.33/14.54 &8.24/13.82/11.10  \\
1e-06 &24.46/45.41/26.20 &24.62/70.16/26.64 &16.48/30.15/19.02 &14.35/24.56/16.95 &11.89/20.54/14.67 &8.23/13.82/11.12  \\
5e-07 & NaN & NaN & NaN & NaN & NaN &8.26/13.86/11.13  \\
\bottomrule
\end{tabular}
\end{adjustbox}
\end{table}

\begin{table}[t]
\caption{
\opt ppl on wikitext/opt/c4 with W4\asym-A8\sym/A8\asym.
See~\tref{tab:weightactivation_quantization_opt_existing_method_full_zqlocal} for all learning rate results of \zqlocal and~\tref{tab:weightactivation_quantization_opt_existing_method_full_zqglobal} of \zqglobal.
}\centering
\label{tab:weightactivation_quantization_opt_existing_method_full_in_appendix}
\begin{adjustbox}{width=0.9\linewidth}
\centering
\begin{tabular}{lcccccccccccccc }
\toprule
Precision     & 125m	& 350m 	& 1.3b	& 2.7b	& 6.7b	& 13b &30b	& 66b \\
\midrule
W4\asym-A8\sym Block\\
\rtn &36.69/44.34/30.60 &26.59/32.13/24.81 &25.31/26.89/22.01 &30.84/35.73/29.01 &164.51/110.85/162.94 &4460.61/3145.51/4255.84 &3216.45/2929.40/3570.19 &3038.22/2930.92/3001.82 \\
\gptq &32.20/38.49/27.47 &24.35/29.82/23.24 &16.28/19.64/16.73 &13.86/17.51/15.00 &46.22/53.98/55.13 &3611.71/2796.71/3820.57 &1738.44/1810.08/2119.82 &5992.87/4115.01/4360.16 \\ 
\zqlocalstar &32.88/38.23/28.20 &25.18/30.06/23.62 &16.78/20.25/17.09 &14.82/18.77/15.61 &16.08/21.15/18.77 &2680.33/1876.48/3052.51 &1884.90/1603.23/1348.08 &575.20/499.42/437.94 \\
\zqglobalstar &32.04/37.48/27.23 &24.01/28.81/22.57 &16.12/19.15/16.23 &13.98/17.70/14.87 &38.27/39.77/52.26 &117.83/141.63/96.83 &253.71/700.40/337.15 &1715.98/1546.50/1799.35\\ 
\midrule 
W4\asym-A8\asym Block \\
\rtn &36.61/44.48/30.64 &25.79/31.28/24.13 &21.23/23.54/19.19 &23.82/29.77/22.60 &13.18/17.04/14.19 &19.87/32.93/26.28 &36.07/136.88/85.84 &627.15/880.79/937.08 \\ 
\gptq &32.22/38.83/27.43 &23.90/29.29/22.63 &15.75/18.74/15.93 &13.23/16.31/14.03 &12.50/15.86/13.29 &12.79/21.99/17.05 &12.96/25.03/24.14 &495.70/681.68/768.69 \\ 
\zqlocalstar &33.60/38.57/28.02 &24.57/29.27/22.98 &15.98/19.13/16.20 &13.44/16.81/14.26 &11.76/14.97/13.00 &11.69/16.98/14.01 &12.38/24.25/18.96 &12.19/23.31/13.47 \\
\zqglobalstar &31.61/37.00/27.10 &23.66/28.56/22.21 &15.77/18.61/15.83 &13.09/16.56/14.00 &12.03/14.60/12.86 &11.80/15.01/12.41 &12.94/17.61/13.41 &31.51/58.00/23.95\\
\midrule
\bottomrule
\end{tabular}
\end{adjustbox}
\end{table}

\begin{table}[t]
\caption{
\opt ppl on wikitext/opt/c4 with W4\asym-A8\sym/A8\asym and \zqlocal. 
}\centering
\label{tab:weightactivation_quantization_opt_existing_method_full_zqlocal}
\begin{adjustbox}{width=0.9\linewidth}
\centering
\begin{tabular}{lcccccccccccccc }
\toprule
Precision     & 125m	& 350m 	& 1.3b	& 2.7b	& 6.7b	& 13b &30b	& 66b \\
\midrule
W4\asym-A8\sym Block\\
0.001 &34.91/40.43/29.37 &26.82/32.68/25.24 &17.68/21.72/18.11 &19.40/27.59/20.05 &36.70/59.32/45.17 &7240.89/5506.67/4889.34 &8229.32/5068.14/5005.13 & Diverge  \\
0.0005 &34.16/39.00/28.58 &26.75/32.05/24.60 &17.19/21.42/17.55 &19.43/25.54/19.41 &29.33/48.38/43.28 &56836.57/36810.64/31073.67 &5448.96/3826.63/3196.49 &575.20/499.42/437.94  \\
0.0001 &32.88/38.23/28.20 &25.31/31.60/23.98 &16.93/20.77/17.36 &17.05/21.50/17.42 &25.24/31.66/26.82 &6125.07/3817.01/4121.70 &1884.90/1603.23/1348.08 &5427.12/3449.58/3289.01  \\
5e-05 &32.86/39.17/27.91 &25.91/31.24/24.07 &16.99/20.02/17.23 &15.07/19.00/15.54 &16.08/21.15/18.77 &6037.51/3617.64/3819.63 &3266.46/2533.64/2463.21 &11631.78/10489.81/7880.43  \\
1e-05 &34.00/39.76/28.62 &25.40/30.60/23.75 &16.87/20.26/17.11 &14.82/18.77/15.61 &26.60/32.09/28.76 &5346.85/3788.29/4903.31 &3364.70/2372.71/3370.97 &5793.44/3544.90/3925.34  \\
5e-06 &34.37/41.46/28.71 &25.18/30.06/23.62 &16.78/20.25/17.09 &14.87/19.42/15.86 &34.53/39.98/38.22 &2680.33/1876.48/3052.51 &3566.45/2532.54/3678.75 &4916.96/3783.69/3716.49  \\
1e-06 &36.05/43.46/30.00 &25.73/30.69/24.05 &19.58/22.57/19.04 &18.66/24.19/19.98 &77.99/62.27/83.19 &3893.00/2672.11/3849.59 &3233.72/2944.44/3732.18 &4238.57/3621.09/3541.33  \\
\midrule 
W4\asym-A8\asym Block \\
0.001 &33.57/40.84/29.00 &27.29/32.48/24.68 &17.41/20.70/17.07 &15.98/20.45/16.23 &12.63/17.21/14.25 &9889.96/7605.54/6328.91 &2009.66/1637.69/2011.15 &5070.07/3124.56/2683.19  \\
0.0005 &34.58/40.45/28.69 &25.81/31.56/24.09 &16.89/20.66/16.93 &15.00/19.47/15.61 &12.55/17.00/14.29 &13.18/19.65/15.18 &36.51/75.89/60.58 &3249.10/63.17/119.55  \\
0.0001 &33.91/38.39/28.12 &25.37/31.24/23.66 &16.78/20.09/16.72 &14.26/18.49/14.90 &12.13/15.97/13.48 &13.48/20.42/16.68 &110.20/117.28/257.96 &12.19/23.31/13.47  \\
5e-05 &33.60/38.57/28.02 &24.67/29.60/23.34 &16.31/19.56/16.42 &13.90/19.16/15.05 &12.30/15.95/13.56 &12.05/18.00/15.77 &37.68/59.83/124.75 &29.72/95.99/69.60  \\
1e-05 &33.80/40.21/28.56 &24.57/29.27/22.98 &15.98/19.13/16.20 &13.44/16.81/14.26 &11.76/14.97/13.00 &11.69/16.98/14.01 &14.39/31.47/24.45 &217.93/313.13/298.24  \\
5e-06 &34.62/41.07/28.93 &24.68/29.46/23.12 &16.26/19.23/16.27 &13.44/17.00/14.36 &11.96/14.86/13.10 &12.31/18.55/15.16 &12.38/24.25/18.96 &85.96/185.07/180.88  \\
1e-06 &35.94/43.35/30.00 &24.92/30.18/23.45 &17.98/20.89/17.45 &14.79/18.90/15.52 &12.10/15.47/13.35 &15.48/22.00/17.84 &14.86/31.16/26.21 &411.89/620.52/652.55  \\
\bottomrule
\end{tabular}
\end{adjustbox}
\end{table}

\begin{table}[t]
\caption{
\opt ppl on wikitext/opt/c4 with W4\asym-A8\sym/A8\asym and \zqglobal. 
}\centering
\label{tab:weightactivation_quantization_opt_existing_method_full_zqglobal}
\begin{adjustbox}{width=0.9\linewidth}
\centering
\begin{tabular}{lcccccccccccccc }
\toprule
Precision     & 125m	& 350m 	& 1.3b	& 2.7b	& 6.7b	& 13b &30b	& 66b \\
\midrule
W4\asym-A8\sym Block\\
0.001 &34.90/44.82/28.27 &8988.08/5862.33/384.69 &nan/nan/nan &18290.16/9784.37/12099.01 &16014.50/8655.69/12304.55 &248961.98/84832.78/104880.55 &56675.05/23709.03/33007.17 &29782.43/20410.10/23559.66  \\
0.0005 &31.78/38.56/27.20 &39.24/54.15/29.76 &10610.96/9438.99/6752.84 &12499.29/8411.26/10677.01 &nan/nan/nan &74731.13/44494.68/29286.49 &51871.73/28548.95/23056.78 &18717.63/11744.97/12903.33  \\
0.0001 &32.04/37.48/27.23 &24.14/29.21/22.47 &17.04/23.64/17.13 &175.67/165.81/162.24 &12305.50/11472.90/10223.89 &16303.04/10731.12/10669.52 &22548.81/12474.28/7405.46 &7926.43/4377.36/4805.98  \\
5e-05 &32.16/37.54/27.27 &24.15/28.87/22.46 &16.02/19.61/16.59 &13.88/20.27/14.79 &5241.10/3284.47/2187.15 &13297.25/7781.85/7467.30 &9542.44/4543.45/5373.00 & NaN  \\
1e-05 &32.57/38.43/27.53 &24.01/28.81/22.57 &16.12/19.15/16.23 &13.98/17.70/14.87 &99.27/118.19/88.74 &529.82/361.44/256.46 &1936.12/1388.68/947.45 &10077.70/9208.34/11462.28  \\
5e-06 &32.83/38.37/27.71 &24.13/29.30/22.68 &16.45/19.64/16.57 &14.42/18.01/15.27 &70.26/62.28/54.47 &373.82/494.33/170.40 &820.90/847.19/543.59 &1867.57/1878.76/4117.49  \\
1e-06 &34.79/41.79/29.30 &24.68/30.01/23.23 &17.90/21.94/18.01 &14.83/18.63/15.70 &38.27/39.77/52.26 &117.83/141.63/96.83 &261.19/844.40/272.04 &1500.51/1275.54/1649.50  \\
5e-07 & NaN & NaN & NaN & NaN & NaN & NaN &253.71/700.40/337.15 &1715.98/1546.50/1799.35  \\
1e-07 & NaN & NaN & NaN & NaN & NaN & NaN &913.95/1117.58/1065.87 &2012.91/1917.48/1817.92  \\
\midrule 
W4\asym-A8\asym Block \\
0.001 &37.89/47.68/30.43 &9023.01/4309.50/1186.96 &12638.86/nan/9164.64 &11285.86/6477.19/nan &12222.01/6933.34/8989.30 &132962.69/73768.05/59268.76 &328993.91/187752.97/163157.59 &48298.52/30548.89/42797.96  \\
0.0005 &32.65/39.86/27.20 &28.46/36.94/24.68 &nan/nan/nan &nan/nan/nan &23287.96/15508.32/16243.28 &22052.30/10852.90/11588.02 &63084.59/39919.41/42499.90 & NaN  \\
0.0001 &31.61/37.00/27.10 &24.64/29.13/22.28 &16.31/19.71/16.44 &43.76/29.11/33.35 &22024.01/13962.04/14130.94 &10171.49/7200.78/7954.12 &18603.08/11639.42/10798.26 &nan/nan/nan  \\
5e-05 &32.21/37.46/27.18 &23.66/28.56/22.21 &16.02/19.02/15.92 &13.48/17.57/14.24 &839.48/213.76/286.05 &1035.13/nan/1472.08 &8085.92/3545.21/4893.07 &nan/nan/nan  \\
1e-05 &32.35/38.21/27.38 &23.59/28.66/22.24 &15.77/18.61/15.83 &13.09/16.56/14.00 &12.09/14.69/12.90 &11.80/15.01/12.41 &13.76/22.87/15.72 &974.58/1557.95/1039.65  \\
5e-06 &32.59/38.49/27.68 &23.62/28.63/22.33 &15.78/18.80/15.95 &13.23/16.65/14.12 &12.03/14.60/12.86 &12.72/16.31/13.20 &12.94/17.61/13.41 &83.35/137.83/128.11  \\
1e-06 &34.68/41.56/29.26 &24.08/29.21/22.68 &16.66/20.03/16.69 &13.30/16.74/14.33 &12.43/15.52/13.36 &12.28/16.13/13.19 &16.00/19.60/14.88 &31.51/58.00/23.95  \\
5e-07 & NaN & NaN & NaN & NaN & NaN & NaN & NaN &31.09/73.23/24.44  \\
1e-07 & NaN & NaN & NaN & NaN & NaN & NaN & NaN &241.81/544.81/505.58  \\
\bottomrule
\end{tabular}
\end{adjustbox}
\end{table}


\begin{table}[t]
\caption{
\bloom ppl on wikitext/opt/c4 with W4\asym-A8\sym/A8\asym.
See~\tref{tab:weightactivation_quantization_bloom_existing_method_full_zqlocal} for all learning rate results of \zqlocal and~\tref{tab:weightactivation_quantization_bloom_existing_method_full_zqglobal} of \zqglobal.
}\centering
\label{tab:weightactivation_quantization_bloom_existing_method_full_in_appendix}
\begin{adjustbox}{width=0.9\linewidth}
\centering
\begin{tabular}{lcccccccccccccc }
\toprule
Precision     & 560m   &1.1b   & 1.7b  & 3b & 7.1b & 176b \\
\midrule
W4\asym-A8\sym Block\\
\rtn  &25.56/47.53/27.31 &24.80/70.99/26.71 &17.36/31.95/19.89 &14.82/25.63/17.47 &12.33/21.62/15.13 &9.12/15.58/14.04 \\
\gptq &24.13/44.79/25.86 &25.69/68.65/27.08 &16.63/30.54/19.12 &14.18/24.42/16.82 &12.04/21.07/14.75 &8.92/15.16/13.56 \\ 
\zqlocalstar &24.45/45.73/26.22 &19.50/52.67/21.73 &16.71/30.23/19.09 &14.37/24.72/16.99 &12.00/20.79/14.78 &8.52/14.29/11.41 \\
\zqglobalstar &23.93/44.31/25.68 &19.71/51.98/21.85 &16.34/29.36/18.82 &14.13/24.34/16.76 &11.84/20.58/14.59 &8.76/14.60/11.68\\ 
\midrule 
W4\asym-A8\asym Block \\
\rtn &25.37/46.99/27.16 &24.08/68.95/26.17 &17.12/31.46/19.67 &14.74/25.38/17.37 &12.22/21.36/15.00 &8.73/15.10/12.83 \\ 
\gptq &24.09/44.29/25.66 &24.50/67.37/26.62 &16.39/29.83/18.91 &14.13/24.47/16.73 &11.91/20.72/14.62 &8.55/14.74/12.31 \\ 
\zqlocalstar &24.29/45.19/26.10 &19.13/52.89/21.63 &16.54/30.11/18.92 &14.32/24.73/16.94 &11.94/20.63/14.68 &8.33/14.01/11.22\\
\zqglobalstar &23.86/44.16/25.62 &19.54/51.72/21.79 &16.23/29.40/18.68 &14.15/24.29/16.72 &11.80/20.37/14.56 &8.62/14.40/11.49\\
\midrule
\bottomrule
\end{tabular}
\end{adjustbox}
\end{table}

\begin{table}[t]
\caption{
\bloom ppl on wikitext/opt/c4 with W4\asym-A8\sym/A8\asym and \zqlocal. 
}\centering
\label{tab:weightactivation_quantization_bloom_existing_method_full_zqlocal}
\begin{adjustbox}{width=0.9\linewidth}
\centering
\begin{tabular}{lcccccccccccccc }
\toprule
Precision     & 560m   &1.1b   & 1.7b  & 3b & 7.1b & 176b \\
\midrule
W4\asym-A8\sym Block\\
0.001 &25.51/47.89/27.15 &19.73/54.63/22.18 &16.96/31.47/19.44 &14.59/25.69/17.32 &12.51/21.85/15.34 &8.62/14.42/11.50  \\
0.0005 &25.18/47.35/26.95 &19.62/53.64/22.03 &16.98/31.75/19.47 &14.52/25.22/17.18 &12.03/21.01/14.82 &8.59/14.38/11.45  \\
0.0001 &24.79/46.37/26.44 &19.50/52.67/21.73 &16.68/30.51/19.18 &14.44/25.12/17.05 &12.00/20.79/14.78 &8.52/14.29/11.41  \\
5e-05 &24.56/46.29/26.34 &23.93/69.17/26.19 &16.71/30.23/19.09 &14.37/24.72/16.99 &12.05/20.92/14.82 &8.55/14.34/11.44  \\
1e-05 &24.45/45.73/26.22 &23.65/66.73/25.80 &16.66/30.69/19.16 &14.40/24.94/17.02 &12.12/21.14/14.86 &8.65/14.97/12.01  \\
5e-06 &24.48/45.66/26.33 &23.87/67.26/25.84 &16.78/30.72/19.23 &14.44/24.91/17.07 &12.15/21.23/14.88 &8.70/15.04/12.37  \\
1e-06 &24.91/46.35/26.72 &24.09/68.13/26.05 &17.03/31.28/19.52 &14.60/25.18/17.24 &12.22/21.31/14.99 &8.91/15.25/13.35  \\
\midrule 
W4\asym-A8\asym Block \\
0.001 &25.26/46.43/26.98 &19.69/54.26/22.14 &16.88/32.16/19.40 &15.15/26.58/17.76 &12.40/22.29/15.28 &8.40/14.06/11.26  \\
0.0005 &24.89/47.99/26.82 &19.54/53.57/21.98 &16.73/31.02/19.29 &14.50/25.52/17.11 &11.94/20.70/14.76 &8.33/14.01/11.22  \\
0.0001 &24.60/45.75/26.44 &19.13/52.89/21.63 &16.54/30.36/19.10 &14.37/24.91/16.93 &11.94/20.63/14.68 &8.35/14.04/11.24  \\
5e-05 &24.41/45.08/26.23 &23.59/67.14/25.79 &16.54/30.11/18.92 &14.29/24.83/16.92 &11.95/20.71/14.71 &8.36/14.10/11.25  \\
1e-05 &24.29/45.19/26.10 &23.35/65.26/25.38 &16.51/30.20/19.00 &14.32/24.73/16.94 &11.97/20.93/14.74 &8.44/14.30/11.45  \\
5e-06 &24.31/45.25/26.15 &23.41/66.18/25.48 &16.63/30.37/19.09 &14.33/24.74/16.96 &12.03/20.95/14.78 &8.52/14.66/11.86  \\
1e-06 &24.76/45.92/26.62 &23.52/66.38/25.66 &16.81/30.71/19.30 &14.53/24.92/17.14 &12.10/21.07/14.87 &8.62/14.92/12.41  \\
\bottomrule
\end{tabular}
\end{adjustbox}
\end{table}

\begin{table}[t]
\caption{
\bloom ppl on wikitext/opt/c4 with W4\asym-A8\sym/A8\asym and \zqglobal. 
}\centering
\label{tab:weightactivation_quantization_bloom_existing_method_full_zqglobal}
\begin{adjustbox}{width=0.9\linewidth}
\centering
\begin{tabular}{lcccccccccccccc }
\toprule
Precision    & 560m   &1.1b   & 1.7b  & 3b & 7.1b & 176b \\
\midrule
W4\asym-A8\sym Block\\
0.001 &174250016.00/201477664.00/1348168.88 &423532.56/906908.06/322995.69 &573201.81/1089364.38/498071.91 &544376.56/696942.56/540949.06 &nan/nan/nan & NaN  \\
0.0005 &70978.52/29214230.00/1151.72 &2880.81/15732.60/309.13 &505479.44/629035.56/29283.36 &140595.53/181082.25/33785.79 &378033.53/789890.00/191543.91 & NaN  \\
0.0001 &24.04/45.38/25.83 &19.44/52.38/21.77 &16.34/29.36/18.82 &14.32/24.74/16.88 &12.12/22.00/14.80 &249.47/26690.76/26.96  \\
5e-05 &23.93/44.31/25.68 &19.71/51.98/21.85 &16.18/29.71/18.71 &14.13/24.34/16.76 &11.84/20.58/14.59 &9.00/15.57/11.61  \\
1e-05 &23.99/44.44/25.77 &22.75/58.31/23.63 &16.28/29.96/18.81 &14.29/24.53/16.87 &11.87/20.57/14.64 &8.76/14.60/11.68  \\
5e-06 &24.14/44.77/25.90 &23.90/64.81/25.29 &16.36/30.03/18.91 &14.32/24.68/16.95 &11.91/20.60/14.71 &9.07/15.12/11.98  \\
1e-06 &24.62/45.70/26.33 &25.55/71.49/27.44 &16.61/30.47/19.17 &14.51/24.91/17.11 &12.06/20.93/14.86 &11.25/19.93/15.76  \\
\midrule 
W4\asym-A8\asym Block \\
0.001 &9059092.00/2932002.50/131873960.00 &499829.19/393190.53/346682.47 &1260531.12/2019747.88/460627.16 &1022130.19/872164.88/679662.62 &nan/nan/nan & NaN  \\
0.0005 &7633.14/378055.53/1032.16 &4271.83/85847.50/1555.66 &87087.04/217513.30/37000.13 &575008.56/814032.50/230285.80 &1212241.00/2389840.25/1504266.50 & NaN  \\
0.0001 &23.96/45.36/25.80 &19.37/52.25/21.88 &16.29/29.36/18.81 &14.32/24.66/16.86 &12.05/22.30/14.77 &1400.84/11880.12/392.79  \\
5e-05 &23.86/44.16/25.62 &19.54/51.72/21.79 &16.23/29.40/18.68 &14.15/24.29/16.72 &11.82/20.44/14.54 &8.73/20.30/11.41  \\
1e-05 &23.96/44.24/25.72 &22.55/58.10/23.49 &16.27/29.82/18.78 &14.16/24.35/16.80 &11.80/20.37/14.56 &8.62/14.40/11.49  \\
5e-06 &24.01/44.68/25.83 &23.67/64.20/25.08 &16.30/29.96/18.85 &14.24/24.49/16.86 &11.81/20.50/14.60 &8.69/14.56/11.58  \\
1e-06 &24.53/45.60/26.26 &24.82/71.17/26.84 &16.55/30.35/19.10 &14.40/24.76/17.01 &11.97/20.83/14.77 &9.14/16.63/17.69  \\
\bottomrule
\end{tabular}
\end{adjustbox}
\end{table}

\begin{table}[t]
\caption{
\opt full results of \tref{tab:opt-4bit-blocksize}.
}\centering
\label{tab:opt-4bit-blocksize-full}
\begin{adjustbox}{width=0.9\linewidth}
\centering
\begin{tabular}{lcccccccccccccc }
\toprule
Method     & 125m	& 350m 	& 1.3b	& 2.7b	& 6.7b	& 13b &30b	& 66b \\
\midrule
BS=1024 \\
\rtn  &N/A &25.42/30.62/23.61 &16.90/19.78/16.59 &N/A &11.63/14.41/12.65 &10.47/13.09/11.75 &9.97/12.40/11.09 &9.83/12.31/10.77 \\
&N/A &26.55 &17.76 &N/A &12.90 &11.77 &11.15 &10.97 \\
\gptq &N/A &23.65/29.09/22.43 &15.16/18.00/15.34 &N/A &11.10/13.40/11.99 &10.28/12.49/11.29 &9.58/11.91/10.75 &9.56/11.61/10.44\\
&N/A &25.05 &16.17 &N/A &12.16 &11.36 &10.75 &10.54\\
\zqglobalstar &N/A &23.27/27.97/21.93 &12.93/15.90/13.64 &N/A &10.98/13.60/12.04 &10.33/12.69/11.50 &9.78/12.16/10.90 &9.52/11.58/10.46\\
&N/A &24.39 &16.18 &N/A &12.21 &11.50 &10.95 &10.52\\
\midrule
BS=512 \\
\rtn &N/A &25.05/29.74/23.21 &15.71/19.05/16.09 &13.67/16.93/14.23 &11.32/14.22/12.50 &10.45/12.99/11.68 &10.03/12.27/11.03 &9.83/12.15/10.67 \\
&N/A &26.00 &16.95 &14.94 &12.68 &11.71 &11.11 &10.89 \\
\gptq &N/A &23.33/28.48/22.13 &15.15/17.95/15.26 &12.65/15.61/13.53 &10.94/13.37/11.94 &10.18/12.49/11.29 &9.58/11.87/10.75 &9.53/11.59/10.43\\
&N/A &24.65 &16.12 &13.93 &12.08 &11.32 &10.73 &10.52\\
\zqglobalstar &N/A &23.41/27.67/21.92 &14.91/17.73/15.25 &12.92/15.59/13.55 &11.08/13.51/11.99 &10.29/12.68/11.46 &9.79/12.16/10.87 &9.51/11.65/10.44\\
&N/A &24.34 &15.97 &14.02 &12.19 &11.48 &10.94 &10.53\\
\midrule
BS=256 \\
\rtn  &31.62/38.19/27.62 &24.76/29.44/22.96 &15.54/18.96/15.90 &13.56/16.62/14.02 &11.19/14.12/12.40 &10.39/12.93/11.61 &9.95/12.24/10.98 &9.70/12.09/10.62\\
&32.48 &25.72 &16.80 &14.73 &12.57 &11.64 &11.06 &10.80\\
\gptq &30.56/37.20/26.68 &23.37/28.33/21.97 &14.95/17.63/15.16 &12.59/15.60/13.49 &10.93/13.29/11.92 &10.15/12.43/11.27 &9.58/11.91/10.74 &9.49/11.60/10.40\\
&31.48 &24.56 &15.91 &13.89 &12.05 &11.28 &10.74 &10.50\\
\zqglobalstar &30.45/35.35/26.24 &23.06/27.72/21.74 &14.93/17.45/15.15 &12.99/15.47/13.50 &10.96/13.45/12.00 &10.25/12.61/11.43 &9.73/12.14/10.89 &9.49/11.58/10.42\\
&30.68 &24.17 &15.84 &13.99 &12.14 &11.43 &10.92 &10.50\\
\midrule
BS=128 \\
\rtn  &30.62/36.67/27.10 &24.12/29.34/22.70 &15.35/18.52/15.66 &13.19/16.24/13.88 &11.11/13.94/12.28 &10.31/12.82/11.54 &9.93/12.12/10.93 &9.56/11.85/10.56\\
&31.47 &25.39 &16.51 &14.43 &12.44 &11.56 &11.00 &10.65\\
\gptq &30.76/36.13/26.52 &23.29/27.94/21.98 &14.93/17.51/15.10 &12.49/15.59/13.46 &10.87/13.34/11.90 &10.11/12.47/11.27 &9.60/11.88/10.73 &9.44/11.53/10.40\\
&31.14 &24.40 &15.85 &13.85 &12.03 &11.28 &10.74 &10.45\\
\zqglobalstar &29.52/34.63/25.98 &22.78/27.56/21.65 &15.02/17.50/15.07 &12.67/15.37/13.45 &10.92/13.42/11.96 &10.16/12.61/11.41 &9.74/12.01/10.82 &9.43/11.49/10.40\\
&30.04 &23.99 &15.86 &13.83 &12.10 &11.39 &10.86 &10.44\\
\midrule
BS=64 \\
\rtn  &30.74/36.68/26.87 &24.28/28.95/22.59 &15.21/18.15/15.47 &13.20/16.13/13.75 &11.01/13.71/12.17 &10.27/12.79/11.49 &9.82/12.05/10.89 &9.46/11.70/10.49\\
&31.43 &25.27 &16.28 &14.36 &12.30 &11.52 &10.92 &10.55\\
\gptq &30.25/35.72/26.43 &23.39/27.55/21.75 &14.81/17.40/15.06 &12.54/15.54/13.44 &10.87/13.29/11.89 &10.09/12.44/11.27 &9.55/11.89/10.72 &9.33/11.49/10.38\\
&30.80 &24.23 &15.76 &13.84 &12.02 &11.27 &10.72 &10.40\\
\zqglobalstar &29.69/34.24/25.72 &22.94/27.49/21.54 &14.90/17.43/15.01 &12.80/15.47/13.44 &10.92/13.33/11.93 &10.21/12.58/11.38 &9.69/12.01/10.81 &9.41/11.49/10.39\\
&29.88 &23.99 &15.78 &13.90 &12.06 &11.39 &10.84 &10.43\\
\midrule
BS=32 \\
\rtn &30.48/36.32/26.64 &23.88/28.66/22.36 &14.99/17.87/15.32 &12.89/16.00/13.67 &10.89/13.70/12.13 &10.32/12.73/11.45 &9.76/12.00/10.85 &9.56/11.55/10.44\\
&31.14 &24.97 &16.06 &14.18 &12.24 &11.50 &10.87 &10.52 \\
\gptq &29.13/34.89/25.90 &23.09/27.59/21.65 &14.80/17.41/15.04 &12.45/15.55/13.42 &10.89/13.32/11.89 &10.08/12.48/11.27 &9.51/11.92/10.73 &Diverge\\
&29.97 &24.11 &15.75 &13.81 &12.03 &11.28 &10.72 &Diverge\\
\zqglobalstar &28.93/34.29/25.63 &22.85/27.23/21.50 &14.80/17.34/14.99 &12.74/15.32/13.40 &10.82/13.36/11.91 &10.23/12.61/11.37 &9.68/11.95/10.80 &9.37/11.47/10.38 \\
&29.62 &23.86 &15.71 &13.82 &12.03 &11.41 &10.81 &10.41\\
\bottomrule
\end{tabular}
\end{adjustbox}
\end{table}

\begin{table}[t]
\caption{
\bloom W4\asym-A16 with various block-size out of the best result from \gptq and \zqglobal.
}\centering
\label{tab:bloom-4bit-blocksize-full}
\begin{adjustbox}{width=0.9\linewidth}
\centering
\begin{tabular}{lcccccccccccccc }
\toprule
Method     & 560m   &1.1b   & 1.7b  & 3b & 7.1b & 176b \\
\midrule
BS=1024 \\
\rtn &24.90/46.37/26.68 &N/A &16.57/30.14/19.00 &N/A &1019.51/1351.45/601.35 &53.41/160.05/43.64\\
&32.65 &N/A &21.90 &N/A &990.77 &85.70\\
\gptq &23.90/43.99/25.47 &N/A &16.12/29.13/18.61 &N/A &11.57/19.82/14.33 &8.16/13.70/11.02\\
&31.12 &N/A &21.29 &N/A &15.24 &10.96\\
\zqglobal &23.62/43.90/25.41 &N/A &15.98/28.67/18.44 &N/A &11.91/20.84/14.58 &8.23/13.94/11.09 \\
&30.98 &N/A &21.03 &N/A &15.78 &11.09\\
\midrule
BS=512 \\
\rtn &24.78/46.07/26.45 &19.41/53.64/21.85 &16.47/29.84/18.88 &14.29/24.84/17.05 &142.38/314.10/100.09 &33.88/103.57/31.02\\
&32.44 &31.63 &21.73 &18.73 &185.52 &56.16\\
\gptq &23.63/43.96/25.36 &18.52/49.73/20.91 &16.07/29.87/18.50 &13.79/23.77/16.41 &11.54/19.75/14.30 &8.14/13.70/11.02\\
&30.98 &29.72 &21.48 &17.99 &15.20 &10.95\\
\zqglobal &23.50/43.53/25.23 &18.31/49.06/20.82 &15.93/28.47/18.38 &13.82/23.92/16.47 &11.85/20.17/14.42 &8.20/13.86/11.07\\
&30.75 &29.40 &20.93 &18.07 &15.48 &11.04\\
\midrule
BS=256 \\
\rtn &24.09/45.13/26.02 &18.87/52.29/21.44 &16.27/29.72/18.76 &14.16/24.42/16.90 &121.09/281.67/88.59 &12.55/27.29/15.60\\
&31.75 &30.87 &21.58 &18.49 &163.78 &18.48\\
\gptq &23.31/43.43/25.12 &18.36/49.13/20.79 &16.07/29.10/18.46 &13.76/23.61/16.38 &11.55/19.72/14.29 &8.14/13.70/11.01\\
&30.62 &29.42 &21.21 &17.92 &15.18 &10.95\\
\zqglobal &23.17/43.16/25.13 &18.24/48.78/20.75 &15.81/28.71/18.32 &13.79/23.69/16.42 &11.59/19.92/14.36 &8.17/13.80/11.06\\
&30.49 &29.26 &20.95 &17.97 &15.29 &11.01\\
\midrule
BS=128 \\
\rtn &23.82/44.78/25.75 &18.62/51.31/21.17 &16.13/29.89/18.66 &14.00/24.19/16.71 &23.90/49.80/24.15 &8.84/15.62/11.70\\
&31.45 &30.37 &21.56 &18.30 &32.62 &12.06\\
\gptq &23.27/43.10/24.99 &18.14/48.72/20.73 &16.03/28.96/18.41 &13.72/23.65/16.34 &11.52/19.73/14.26 &8.14/13.67/11.01\\
&30.45 &29.20 &21.13 &17.90 &15.17 &10.94\\
\zqglobal &23.14/42.95/24.97 &18.17/48.53/20.70 &15.75/28.71/18.29 &13.73/23.65/16.37 &11.56/19.77/14.32 &8.17/13.78/11.03\\
&30.35 &29.13 &20.92 &17.92 &15.22 &10.99\\
\midrule
BS=64 \\
\rtn &23.65/44.04/25.51 &18.53/50.02/21.03 &16.06/29.57/18.60 &13.93/23.95/16.60 &11.85/20.51/14.65 &8.31/14.14/11.18\\
&31.07 &29.86 &21.41 &18.16 &15.67 &11.21\\
\gptq &23.11/42.95/24.94 &18.14/48.87/20.65 &16.00/28.91/18.38 &13.72/23.68/16.33 &11.51/19.70/14.27 &8.14/13.69/11.00\\
&30.33 &29.22 &21.10 &17.91 &15.16 &10.94\\
\zqglobal &23.00/42.80/24.91 &18.10/48.30/20.64 &15.68/28.55/18.25 &13.70/23.63/16.36 &11.53/19.67/14.27 &8.17/13.72/11.02\\
&30.24 &29.01 &20.82 &17.90 &15.16 &10.97\\
\midrule
BS=32 \\
\rtn &23.60/43.91/25.50 &18.63/50.13/21.04 &15.98/29.56/18.56 &13.92/23.90/16.53 &11.65/20.01/14.43 &8.20/13.86/11.07\\
&31.00 &29.93 &21.37 &18.12 &15.36 &11.04\\
\gptq &23.10/43.19/24.91 &18.17/48.35/20.66 &15.95/28.95/18.36 &13.76/23.60/16.33 &11.53/19.71/14.27 &8.14/13.70/11.00\\
&30.40 &29.06 &21.08 &17.89 &15.17 &10.95\\
\zqglobal &23.07/42.63/24.82 &18.07/48.07/20.59 &15.66/28.58/18.21 &13.72/23.59/16.33 &11.52/19.71/14.26 &8.16/13.69/11.01\\
&30.18 &28.91 &20.82 &17.88 &15.16 &10.95\\
\bottomrule
\end{tabular}
\end{adjustbox}
\end{table}

\begin{table}[t]
\caption{
\opt full results of three-bit weight with various block-size.
}\centering
\label{tab:opt-3bit-blocksize-full}
\begin{adjustbox}{width=0.9\linewidth}
\centering
\begin{tabular}{lcccccccccccccc }
\toprule
Method     & 125m	& 350m 	& 1.3b	& 2.7b	& 6.7b	& 13b &30b	& 66b \\
\midrule
Full Row \\
\rtn  &2095.20/1848.83/1222.00 &47.43/53.38/36.93 &4399.18/4400.98/3551.88 &8326.78/4208.57/4895.83 &878.00/735.86/910.10 &1953.43/1953.60/1669.76 &439.39/691.94/437.96 &1465.06/1564.59/1282.58\\
&1722.01 &45.91 &4117.35 &5810.40 &841.32 &1858.93 &523.09 &1437.41\\
\gptq &845.81/599.71/496.14 &30.65/34.09/26.15 &20.23/27.39/19.45 &15.91/19.26/16.01 &12.69/15.90/13.96 &11.36/13.71/12.21 &10.10/12.54/11.20 &16.77/21.16/15.39\\
&647.22 &30.30 &22.36 &17.06 &14.18 &12.43 &11.28 &17.77\\
\zqglobalstar &46.47/58.55/35.45 &29.64/36.51/25.55 &32.48/94.57/28.97 &60.91/116.22/36.45 &23.87/29.75/23.88 &44.70/60.78/46.18 &13.16/20.49/13.48 &28.93/75.91/27.28\\
&46.82 &30.57 &52.01 &71.19 &25.83 &50.55 &15.71 &44.04\\
\midrule
BS=1024 \\
\rtn &N/A &44.57/49.58/35.09 &1950.00/2317.55/1913.55 &3810.79/2563.06/3054.91 &50.01/70.17/99.21 &265.62/417.03/261.93 &362.47/252.33/364.45 &523.81/846.60/1021.17\\
&N/A &43.08 &2060.37 &3142.92 &73.13 &314.86 &326.42 &797.20\\
\gptq &N/A &29.78/33.76/25.66 &19.03/23.32/18.14 &N/A &11.69/14.31/12.70 &10.56/12.96/11.70 &9.89/12.19/11.02 &12.84/16.17/13.02\\
&N/A &29.73 &20.16 &N/A &12.90 &11.74 &11.03 &14.01\\
\zqglobalstar &N/A &29.19/34.57/25.11 &19.83/29.77/19.79 &N/A &13.99/18.82/14.76 &13.43/19.28/13.76 &11.10/14.46/11.94 &11.87/14.86/12.13\\
&N/A &29.62 &23.13 &N/A &15.86 &15.49 &12.50 &12.95\\
\midrule
BS=512 \\
\rtn &N/A &37.74/45.10/31.85 &1777.53/1304.55/852.03 &1604.07/1407.49/1487.78 &25.13/40.56/40.08 &130.75/175.33/135.67 &620.53/340.68/416.28 &198.01/457.78/426.15\\
&N/A &38.23 &1311.37 &1499.78 &35.26 &147.25 &459.16 &360.65\\
\gptq &N/A &28.46/32.54/25.14 &18.02/21.35/17.46 &14.38/17.24/14.79 &11.57/14.33/12.57 &10.41/12.97/11.64 &9.77/12.18/10.97 &11.89/14.48/12.40\\
&N/A &28.71 &18.94 &15.47 &12.82 &11.67 &10.97 &12.92\\
\zqglobalstar &N/A &27.81/33.57/24.55 &18.31/23.54/17.99 &18.10/29.47/17.15 &12.54/16.60/13.62 &11.82/15.98/12.81 &10.48/13.36/11.66 &11.26/13.95/11.79\\
&N/A &28.65 &19.95 &21.57 &14.25 &13.54 &11.83 &12.33\\
\midrule
BS=256 \\
\rtn &4349.14/2907.61/2510.75 &35.36/42.07/30.81 &127.17/358.19/142.49 &670.51/550.66/531.80 &19.10/32.39/27.26 &42.52/56.35/43.32 &32.84/60.38/33.48 &210.01/478.13/413.00\\
&3255.84 &36.08 &209.28 &584.32 &26.25 &47.40 &42.23 &367.05\\
\gptq &41.81/49.95/32.48 &27.60/33.73/24.88 &16.97/20.19/16.70 &13.69/17.06/14.54 &11.65/14.24/12.48 &10.35/12.93/11.61 &9.66/12.10/10.93 &11.60/13.98/11.92\\
&41.41 &28.74 &17.95 &15.10 &12.79 &11.63 &10.90 &12.50\\
\zqglobalstar &38.60/46.57/31.36 &26.88/32.79/24.08 &16.82/21.21/17.05 &14.86/19.63/15.37 &11.86/15.87/13.10 &11.33/14.95/12.48 &10.41/12.95/11.41 &10.26/12.66/11.08\\
&38.85 &27.92 &18.36 &16.62 &13.61 &12.92 &11.59 &11.34\\
\midrule
BS=128 \\
\rtn &3446.89/2156.26/1484.15 &33.13/41.23/29.51 &49.40/88.45/45.07 &153.68/155.21/113.98 &16.34/26.86/21.98 &17.80/25.95/18.28 &45.83/43.91/57.50 &106.84/241.02/212.94\\
&2362.43 &34.62 &60.97 &140.96 &21.72 &20.67 &49.08 &186.93 \\
\gptq &40.00/45.73/31.15 &27.68/34.04/25.18 &16.47/19.90/16.47 &13.81/16.96/14.37 &11.57/14.10/12.41 &10.35/12.84/11.58 &9.73/12.08/10.91 &10.96/13.27/11.45\\
&38.96 &28.97 &17.61 &15.05 &12.69 &11.59 &10.91 &11.90\\
\zqglobalstar &36.57/43.88/29.94 &25.75/31.59/23.57 &16.28/20.20/16.67 &14.27/18.41/14.90 &11.70/15.05/12.68 &11.13/15.07/12.17 &10.31/12.99/11.32 &10.12/12.66/11.01\\
&36.80 &26.97 &17.72 &15.86 &13.14 &12.79 &11.54 &11.27\\
\midrule
BS=64 \\
\rtn &708.02/477.13/287.03 &32.61/42.14/29.09 &25.43/38.84/24.63 &72.84/69.27/48.07 &14.11/21.71/16.56 &14.13/20.08/15.25 &20.55/32.74/24.49 &30.66/70.73/65.57\\
&490.73 &34.61 &29.63 &63.39 &17.46 &16.48 &25.93 &55.65\\
\gptq &37.15/42.59/30.07 &27.68/33.55/25.12 &16.25/19.80/16.32 &13.66/16.69/14.37 &11.42/13.98/12.37 &10.37/12.90/11.58 &9.68/12.17/10.92 &10.39/12.65/11.15\\
&36.60 &28.78 &17.46 &14.91 &12.59 &11.62 &10.92 &11.40\\
\zqglobalstar &35.82/40.98/29.65 &25.31/31.60/23.38 &16.05/19.77/16.39 &13.33/16.92/14.31 &11.56/14.70/12.59 &10.88/13.64/12.04 &10.04/12.70/11.27 &10.04/12.06/10.81\\
&35.48 &26.76 &17.40 &14.85 &12.95 &12.19 &11.34 &10.97\\
\midrule
BS=32 \\
\rtn &72.83/88.62/54.25 &32.36/40.76/29.06 &20.22/27.31/19.81 &31.12/42.01/26.83 &13.38/18.56/15.44 &13.06/18.35/14.38 &11.12/15.05/12.35 &19.29/43.61/34.10\\
&71.90 &34.06 &22.44 &33.32 &15.79 &15.26 &12.84 &32.33\\
\gptq &38.26/45.01/30.92 &27.16/33.65/24.97 &16.13/19.83/16.45 &13.66/17.06/14.50 &11.43/14.08/12.42 &10.48/12.96/11.65 &9.78/12.24/10.96 &Diverge\\
&38.06 &28.59 &17.47 &15.07 &12.64 &11.70 &10.99 &Diverge\\
\zqglobalstar &33.44/39.48/28.33 &25.19/30.73/23.22 &15.62/19.52/16.20 &13.35/16.64/14.18 &11.56/14.38/12.61 &10.86/13.64/12.03 &10.25/12.86/11.28 &9.99/12.05/10.81\\
&33.75 &26.38 &17.11 &14.73 &12.85 &12.17 &11.46 &10.95\\
\bottomrule
\end{tabular}
\end{adjustbox}
\end{table}

\begin{table}[t]
\caption{
\bloom W3\asym-A16 with various block-size out of the best result from \gptq and \zqglobal.
}\centering
\label{tab:bloom-3bit-blocksize-full}
\begin{adjustbox}{width=0.9\linewidth}
\centering
\begin{tabular}{lcccccccccccccc }
\toprule
Method     & 560m   &1.1b   & 1.7b  & 3b & 7.1b & 176b \\
Full row \\
\rtn &68.45/132.83/59.22 &118.61/317.41/99.65 &31.15/67.23/34.02 &31.07/59.03/32.17 &66140.72/78568.16/44504.19 &100371.84/166012.19/137892.34 \\
&86.83 &178.56 &44.14 &40.76 &63071.02 &134758.79\\
\gptq &46.92/84.69/39.50 &49.78/142.95/43.84 &19.70/41.35/21.74 &22.84/46.49/22.90 &52966.59/52979.88/37115.48 &Diverge\\
&57.04 &78.85 &27.59 &30.74 &47687.32 &Diverge\\
\zqglobal &33.20/64.61/32.30 &34.16/100.05/29.22 &19.22/36.30/21.25 &18.41/33.10/20.79 &273.55/439.59/100.79 &27.19/75.74/45.45\\
&43.37 &54.48 &25.59 &24.10 &271.31 &49.46\\
\midrule
\midrule
BS=1024 \\
\rtn &47.00/86.57/43.37 &70.81/230.74/70.78 &35.41/65.75/33.54 &22.12/40.65/24.55 &25654.77/25531.66/15868.46 &141324.41/183583.73/200436.33\\
&58.98 &124.11 &44.90 &29.11 &22351.63 &175114.82\\
\gptq &31.25/58.80/30.94 &N/A &19.11/37.07/20.90 &N/A &12.59/21.95/15.21 &8.31/13.96/11.17 \\
&40.33 &N/A &25.69 &N/A &16.58 &11.15 \\
\zqglobal &28.91/55.81/29.59 &N/A &18.20/34.13/20.40 &N/A &30.94/119.98/21.39 &15.98/32.85/19.85\\
&38.10 &N/A &24.24 &N/A &57.44 &22.89\\
\midrule
BS=512 \\
\rtn &41.58/79.83/39.41 &33.83/116.88/37.34 &25.95/49.65/26.77 &19.94/38.58/22.58 &9777.49/8000.29/5407.46 &202051.34/273707.81/279776.97\\
&53.61 &62.68 &34.12 &27.03 &7728.41 &251845.38\\
\gptq &28.08/53.15/29.05 &21.20/61.42/23.33 &18.41/34.47/20.43 &15.08/26.14/17.53 &12.32/21.29/15.01 &8.30/13.98/11.16\\
&36.76 &35.32 &24.44 &19.58 &16.21 &11.15\\
\zqglobal &26.80/50.49/28.31 &20.77/57.57/22.89 &17.64/33.19/19.91 &15.16/26.51/17.57 &16.35/28.75/15.76 &11.38/20.36/14.66\\
&35.20 &33.75 &23.58 &19.75 &20.29 &15.47\\
\midrule
BS=256 \\
\rtn &36.13/70.37/36.29 &28.65/95.72/31.80 &21.67/42.59/23.80 &17.64/32.82/20.69 &1322.61/1864.55/946.92 &166006.80/187829.98/198052.83\\
&47.60 &52.06 &29.35 &23.72 &1378.02 &183963.20\\
\gptq &27.10/51.11/28.24 &20.60/56.57/22.77 &17.97/33.28/20.04 &14.82/25.79/17.31 &12.27/21.24/14.93 &8.27/13.99/11.14\\
&35.48 &33.31 &23.76 &19.31 &16.15 &11.13\\
\zqglobal &25.96/49.75/27.59 &20.21/54.83/22.33 &17.43/32.14/19.67 &14.85/25.79/17.33 &12.85/22.00/15.04 &9.07/15.88/11.88\\
&34.43 &32.46 &23.08 &19.32 &16.63 &12.28\\
\midrule
BS=128 \\
\rtn &34.71/66.56/35.27 &24.43/73.77/26.90 &19.59/37.22/21.98 &16.11/28.81/18.89 &108.32/252.15/74.42 &111057.84/101926.99/105339.26\\
&45.51 &41.70 &26.26 &21.27 &144.96 &106108.03\\
\gptq &26.29/49.86/27.54 &20.26/55.76/22.42 &17.77/32.65/19.92 &14.58/25.25/17.11 &12.18/21.06/14.86 &8.26/13.92/11.12\\
&34.56 &32.81 &23.45 &18.98 &16.03 &11.10\\
\zqglobal &25.28/48.24/26.96 &19.79/54.04/22.03 &17.12/31.42/19.31 &14.62/25.73/17.17 &12.04/21.02/14.82 &8.43/14.44/11.29\\
&33.49 &31.95 &22.62 &19.17 &15.96 &11.39\\
\midrule
BS=64 \\
\rtn &30.88/59.01/32.08 &23.04/67.93/25.49 &19.35/37.67/21.80 &15.64/27.56/18.39 &37.15/65.22/33.22 &198.66/488.11/128.62\\
&40.66 &38.82 &26.27 &20.53 &45.20 &271.80\\
\gptq &26.31/49.91/27.17 &20.11/55.06/22.23 &17.94/32.42/19.76 &14.62/25.39/17.07 &12.13/21.07/14.83 &8.26/13.93/11.11\\
&34.46 &32.47 &23.37 &19.02 &16.01 &11.10\\
\zqglobal &25.17/48.01/26.59 &19.51/53.27/21.75 &16.88/31.14/19.22 &14.51/25.18/17.05 &12.00/20.85/14.74 &8.35/14.06/11.20\\
&33.26 &31.51 &22.41 &18.91 &15.86 &11.21\\
\midrule
BS=32 \\
\rtn &30.15/57.55/31.51 &23.49/70.15/25.56 &18.96/36.54/21.42 &15.56/27.48/18.32 &13.06/23.77/16.05 &10.28/18.90/13.27\\
&39.74 &39.73 &25.64 &20.46 &17.62 &14.15\\
\gptq &25.96/49.99/27.06 &19.97/54.79/22.16 &17.60/32.24/19.76 &14.55/25.76/17.06 &12.20/21.01/14.85 &8.28/13.95/11.13\\
&34.33 &32.31 &23.20 &19.12 &16.02 &11.12\\
\zqglobal &25.09/47.36/26.34 &19.43/52.95/21.64 &16.86/30.49/19.11 &14.50/25.36/16.99 &12.00/20.84/14.72 &8.35/14.04/11.20\\
&32.93 &31.34 &22.15 &18.95 &15.85 &11.20\\
\bottomrule
\end{tabular}
\end{adjustbox}
\end{table}

\begin{table}[t]
\caption{
Full results of \bloom-176B with different quantization bits
}\centering
\label{tab:bloom-176-different-bits-full}
\begin{adjustbox}{width=0.9\linewidth}
\centering
\begin{tabular}{lcccccccccccccc }
\toprule
Bits     & 3   &4    &5 &6 &7 &8 \\
\midrule
Per-row &27.19/75.74/45.45 &8.16/13.70/11.02 &8.13/13.67/10.99 &8.11/13.63/10.98 &8.11/13.62/10.97 &8.10/13.62/10.98\\
1024    &8.31/13.96/11.17  &8.14/13.70/11.02 &8.11/13.62/10.97  &8.11/13.62/10.97 &8.11/13.63/10.97 &N/A\\
64      &8.26/13.93/11.11  &8.14/13.69/11.00 &8.11/13.62/10.96 &N/A &N/A &N/A\\
\bottomrule
\end{tabular}
\end{adjustbox}
\end{table}
\begin{table}[t]
\caption{
\opt full results of \tref{tab:opt-4bit8bit-blocksize}.
}\centering
\label{tab:opt-4bit8bit-blocksize-full}
\begin{adjustbox}{width=0.9\linewidth}
\centering
\begin{tabular}{lcccccccccccccc }
\toprule
Method     & 125m	& 350m 	& 1.3b	& 2.7b	& 6.7b	& 13b &30b	& 66b \\
\midrule
W4\asym full row and A8\sym 128\\
\rtn &36.64/44.84/30.90 &25.58/31.06/23.99 &19.96/22.31/18.20 &18.42/23.01/18.56 &12.04/15.92/13.20 &10.79/13.65/12.11 &10.10/13.17/11.37 &20.50/45.58/25.37\\
&37.46 &26.88 &20.16 &20.00 &13.72 &12.18 &11.54 &30.48\\
\gptq &31.82/38.82/27.54 &23.78/28.96/22.61 &15.56/18.27/15.62 &13.02/15.88/13.76 &11.22/13.59/12.11 &10.25/12.65/11.37 &9.56/11.94/10.79 &9.62/11.72/10.54\\
&32.73 &25.12 &16.48 &14.22 &12.31 &11.42 &10.76 &10.63\\
\zqlocal &&&&& &&&9.79/11.94/10.65\\
&&&&&&&&10.79\\
\zqglobal &31.69/36.66/27.19 &23.47/28.18/22.03 &15.53/18.35/15.73 &13.02/16.11/13.82 &11.29/13.70/12.19 &10.43/12.91/11.64 &9.86/12.28/11.00 &9.62/11.84/10.63\\
&31.85 &24.56 &16.54 &14.32 &12.39 &11.66 &11.05 &10.70\\
\midrule
W4\asym 128 and A8\sym 128\\
\rtn &30.61/36.57/27.08 &24.14/29.47/22.80 &15.46/18.68/15.77 &13.24/16.36/13.95 &11.16/14.08/12.35 &10.35/12.89/11.57 &9.95/12.15/10.95 &9.58/11.90/10.58\\
&31.42 &25.47 &16.64 &14.52 &12.53 &11.60 &11.02 &10.69\\
\gptq &30.47/36.45/26.45 &23.43/28.12/22.06 &14.90/17.62/15.17 &12.51/15.63/13.48 &10.88/13.35/11.93 &10.17/12.48/11.28 &9.58/11.86/10.74 &9.35/11.54/10.40\\
&31.12 &24.54 &15.90 &13.87 &12.05 &11.31 &10.73 &10.43\\
\zqlocal &&&&&&&&9.40/11.63/10.51\\
&&&&&&&&10.51\\
\zqglobal &29.59/34.68/25.91 &22.59/27.93/21.68 &14.87/17.55/15.11 &12.65/15.45/13.48 &10.88/13.40/11.94 &10.20/12.67/11.43 &9.74/12.03/10.83 &9.40/11.51/10.42\\
&30.06 &24.07 &15.84 &13.86 &12.08 &11.43 &10.87 &10.44\\
\midrule
W4\asym full row and A8\asym 128\\
\rtn &36.61/44.71/30.85 &25.50/30.93/23.88 &19.58/22.08/18.01 &19.53/24.38/19.68 &11.91/15.35/13.01 &10.68/13.50/12.02 &10.13/13.21/11.37 &17.90/32.15/20.02\\
&37.39 &26.77 &19.89 &21.20 &13.42 &12.07 &11.57 &23.36\\
\gptq &32.15/39.58/27.65 &23.48/28.92/22.46 &15.43/18.24/15.55 &12.92/15.94/13.74 &11.17/13.59/12.09 &10.35/12.63/11.36 &9.65/11.95/10.79 &9.58/11.71/10.55\\
&33.13 &24.95 &16.40 &14.20 &12.29 &11.45 &10.80 &10.61\\
\zqlocal &&&&&&& &10.05/11.91/10.61\\
&&&&&&&&10.86\\
\zqglobal &31.55/37.49/27.25 &23.34/28.33/22.08 &15.52/18.55/15.61 &13.07/16.09/13.82 &11.32/13.65/12.16 &10.42/12.86/11.63 &9.86/12.30/11.00 &9.67/12.22/10.86\\
&32.10 &24.58 &16.56 &14.33 &12.37 &11.64 &11.05 &10.91\\
\midrule
W4\asym 128 and A8\asym 128\\
\rtn &30.59/36.56/27.07 &24.11/29.43/22.74 &15.38/18.57/15.69 &13.22/16.32/13.91 &11.13/13.97/12.30 &10.34/12.82/11.55 &9.98/12.15/10.96 &9.57/11.86/10.58\\
&31.41 &25.43 &16.55 &14.49 &12.47 &11.57 &11.03 &10.67\\
\gptq &30.47/36.19/26.40 &23.35/27.96/21.94 &14.92/17.57/15.12 &12.48/15.60/13.46 &10.87/13.34/11.91 &10.20/12.45/11.28 &9.62/11.88/10.74 &9.39/11.55/10.41\\
&31.02 &24.42 &15.87 &13.85 &12.04 &11.31 &10.75 &10.45\\
\zqlocal &&&&&&& &9.37/11.70/10.49\\
&&&&&&&&10.52\\
\zqglobal &29.85/34.52/26.10 &22.70/27.72/21.64 &14.96/17.55/15.09 &12.64/15.40/13.47 &10.93/13.43/11.95 &10.18/12.68/11.42 &9.74/12.02/10.83 &9.39/11.53/10.42\\
&30.16 &24.02 &15.86 &13.84 &12.10 &11.42 &10.86 &10.45\\
\bottomrule
\end{tabular}
\end{adjustbox}
\end{table}

\begin{table}[t]
\caption{
\bloom full results of \tref{tab:bloom-176-different-blocks}.
}\centering
\label{tab:bloom-4bit8bit-blocksize-full}
\begin{adjustbox}{width=0.9\linewidth}
\centering
\begin{tabular}{lcccccccccccccc }
\toprule
Method     & 560m   &1.1b   & 1.7b  & 3b & 7.1b & 176b \\
\midrule
W4\asym full row and A8\sym 128\\
\rtn &25.32/46.98/27.12 &23.87/68.29/25.97 &16.99/31.15/19.51 &14.69/25.22/17.30 &12.07/20.86/14.84 &8.34/14.05/11.24\\
&33.14 &39.38 &22.55 &19.07 &15.92 &11.21\\
\gptq &24.00/44.47/25.66 &24.14/66.95/26.17 &16.38/29.64/18.79 &14.10/24.19/16.67 &11.77/20.22/14.48 &8.20/13.82/11.07\\
&31.37 &39.09 &21.61 &18.32 &15.49 &11.03\\
\zqlocal &&&&&&8.30/14.01/11.20 \\
 &&&&&&11.17\\
\zqglobal &23.92/44.23/25.69 &22.53/57.71/23.51 &16.25/29.72/18.74 &14.12/24.26/16.74 &11.78/20.30/14.53 &8.24/13.82/11.10\\
&31.28 &34.58 &21.57 &18.38 &15.53 &11.05\\
\midrule
W4\asym 128 and A8\sym 128\\
\rtn &23.84/44.94/25.79 &18.65/51.54/21.21 &16.18/30.03/18.70 &14.04/24.32/16.77 &23.05/48.33/23.69 &8.87/15.68/11.72\\
&31.53 &30.46 &21.64 &18.38 &31.69 &12.09\\
\gptq &23.22/43.24/25.01 &18.25/48.89/20.74 &16.00/29.44/18.41 &13.77/23.68/16.35 &11.54/19.76/14.27 &8.13/13.69/11.01\\
&30.49 &29.29 &21.29 &17.93 &15.19 &10.95\\
\zqlocal &&&&&&8.20/13.87/11.08\\
&&&&& &11.05\\
\zqglobal &23.12/43.22/25.03 &18.19/48.96/20.72 
&15.75/28.81/18.30 &13.73/23.65/16.39 &11.57/19.85/14.32 &8.17/13.76/11.03\\
&30.45 &29.29 &20.95 &17.92 &15.25 &10.99\\
\midrule
W4\asym full row and A8\asym 128\\
\rtn &25.30/46.87/27.10 &23.90/68.31/25.98 &16.96/31.09/19.48 &14.68/25.19/17.28 &12.07/20.86/14.84 &8.34/14.06/11.24\\
&33.09 &39.39 &22.51 &19.05 &15.92 &11.21\\
\gptq &23.97/44.15/25.62 &24.61/68.19/26.53 &16.36/29.77/18.81 &14.10/24.17/16.66 &11.78/20.32/14.49 &8.20/13.82/11.07\\
&31.24 &39.78 &21.65 &18.31 &15.53 &11.03\\
\zqlocal &&&&& &8.32/13.97/11.20\\
&&&&& &11.16\\
\zqglobal &23.88/44.40/25.68 &22.63/57.91/23.39 &16.25/29.77/18.74 &14.17/24.24/16.74 &11.77/20.28/14.52 &8.25/13.82/11.10\\
&31.32 &34.64 &21.59 &18.38 &15.52 &11.06\\
\midrule
W4\asym 128 and A8\asym 128\\
\rtn &23.83/44.89/25.77 &18.63/51.46/21.19 &16.16/29.95/18.68 &14.03/24.27/16.75 &23.51/49.07/23.96 &8.85/15.65/11.72\\
&31.50 &30.43 &21.60 &18.35 &32.18 &12.08\\
\gptq &23.26/43.24/25.00 &18.18/48.84/20.73 &16.05/29.34/18.42 &13.69/23.56/16.34 &11.54/19.75/14.28 &8.14/13.71/11.02\\
&30.50 &29.25 &21.27 &17.86 &15.19 &10.96\\
\zqlocal &&&&&&8.19/13.90/11.07\\
&&&&&&11.06\\
\zqglobal &23.12/43.14/25.01 &18.18/48.99/20.73 &15.71/28.73/18.30 &13.74/23.68/16.39 &11.56/19.85/14.31 &8.17/13.78/11.04\\
&30.42 &29.30 &20.91 &17.94 &15.24 &11.00\\
\bottomrule
\end{tabular}
\end{adjustbox}
\end{table}

\begin{table}[t]
\caption{
Full results of~\tref{tab:bloom-176-different-blocks}.
}\centering
\label{tab:bloom-176-different-blocks-full}
\begin{adjustbox}{width=0.9\linewidth}
\centering
\begin{tabular}{lcccccccccccccc }
\toprule
Block SIze     & 1024   &512    &256 &128 &64 &32 \\
\midrule
\ppl &8.16/13.75/11.04 &8.15/13.75/11.02 &8.15/13.70/11.01 &8.13/13.69/11.01 &8.14/13.69/11.01 &8.14/13.69/11.01\\
\bottomrule
\end{tabular}
\end{adjustbox}
\end{table}

\begin{table}[t]
\caption{
Results of applying \lorc on top of \zqglobal for INT8 Activation.
}\centering
\label{tab:LORC-int8}
\begin{adjustbox}{width=0.9\linewidth}
\centering
\begin{tabular}{lcc|ccccc|ccccccc }
\toprule
                      &                    &            & \multicolumn{5}{c|}{Learning Rate}                    &        \\
model-size            & precision         & LoRC-dim & 0.0005   & 0.0001   & 5.00E-05 & 1.00E-05 & 5.00E-06 & Best   \\\midrule
\multirow{3}{*}{125m} & \multirow{3}{*}{W4A8} & 0          & 4482.1   & 31.15    & 30.40     & 30.55    & 30.72    & 30.40   \\
                      &                    & 8          & 5996.14  & 30.96    & 30.24    & 30.37    & 30.61    & 30.24  \\
                      &                    & 16         & 3577.12  & 31.02    & 30.26    & 30.2     & 30.37    & 30.20   \\\midrule
 \multirow{3}{*}{125m}& \multirow{3}{*}{W3A8} & 0          & 4283.28  & 41.03    & 40.93    & 55.74    & 86.34    & 40.93  \\
                      &                    & 8          & 2396.92  & 37.25    & 36.65    & 37.85    & 39.06    & 36.65  \\
                      &                    & 16         & 1787.74  & 36.66    & 36.55    & 37.46    & 38.21    & 36.55  \\\midrule
\multirow{3}{*}{125m}& \multirow{3}{*}{W2A8} & 0          & 3473.18  & 583.72   & 996.76   & 2480.69  & 3203.11  & 583.72 \\
                      &                    & 8          & 3815.37  & 144.85   & 160.71   & 362.17   & 466.98   & 144.85 \\
                      &                    & 16         & 3324.23  & 135.25   & 156.28   & 295.78   & 372.7    & 135.25 \\\toprule
                                         &                    &            & \multicolumn{5}{c}{Learning Rate}                    &        \\
                      &                    & LoRC-dim & 5.00E-05 & 1.00E-05 & 5.00E-06 & 1.00E-06 & 5.00E-07 & best   \\\midrule
\multirow{3}{*}{350m} & \multirow{3}{*}{W4A8} & 0          & 25.65    & 24.38    & 24.34    & 24.55    & 24.75    & 24.34  \\
                      &                    & 8          & 25.56    & 24.3     & 24.24    & 24.45    & 24.66    & 24.24  \\
                      &                    & 16         & 25.45    & 24.39    & 24.21    & 24.39    & 24.63    & 24.21  \\\midrule
  \multirow{3}{*}{350m}                     & \multirow{3}{*}{W3A8} & 0          & 30.59    & 28.45    & 28.94    & 31.51    & 32.39    & 28.45  \\
                      &                    & 8          & 30.1     & 28.22    & 28.71    & 30.81    & 32.09    & 28.22  \\
                      &                    & 16         & 30.64    & 28.02    & 28.50     & 30.62    & 31.69    & 28.02  \\\midrule
  \multirow{3}{*}{350m}                     & \multirow{3}{*}{W2A8} & 0          & 97.40     & 177.43   & 257.61   & 668.19   & 722.19   & 97.4   \\
                      &                    & 8          & 95.79    & 139.68   & 194.36   & 437.18   & 459.92   & 95.79  \\
                      &                    & 16         & 106.51   & 137.81   & 172.93   & 400.91   & 421.59   & 106.51\\
\bottomrule
\end{tabular}
\end{adjustbox}
\end{table}


\end{document}